%% file: arxiv_2024.tex
\documentclass{article}

\usepackage[preprint,nonatbib]{arxiv_2024}

\usepackage[utf8]{inputenc} % allow utf-8 input
\usepackage[T1]{fontenc}    % use 8-bit T1 fonts
\usepackage{hyperref}       % hyperlinks
\usepackage{url}            % simple URL typesetting
\usepackage{booktabs}       % professional-quality tables
\usepackage{amsfonts}       % blackboard math symbols
\usepackage{nicefrac}       % compact symbols for 1/2, etc.
\usepackage{microtype}      % microtypography
\usepackage{xcolor}         % colors

\usepackage{graphicx}
\usepackage{subfigure}
\usepackage{booktabs} % for professional tables
\usepackage{array}
\usepackage{multirow}
\usepackage{makecell}
\usepackage{amsmath}
\usepackage{siunitx}
\usepackage{float}
\usepackage{enumitem}
\usepackage{times}
\usepackage{epsfig}
\usepackage{adjustbox}
\usepackage{stfloats}
\usepackage{wrapfig}
\usepackage{lineno}
\usepackage{tabularx}

\usepackage{comment}

\title{Accuracy Booster: Enabling 4-bit Fixed-point Arithmetic for DNN Training}

\author{%
  \begin{tabularx}{\textwidth}{>{\centering\arraybackslash\normalfont}X>{\centering\arraybackslash\normalfont}X>{\centering\arraybackslash\normalfont}X}
    \textbf{Simla Burcu Harma} & \textbf{Ayan Chakraborty} & \textbf{Nicholas Sperry} \\
    \small{EcoCloud, EPFL} & \small{EcoCloud, EPFL} & \small{EPFL} \\
    \small{\texttt{\href{mailto:simla.harma@epfl.ch}{simla.harma@epfl.ch}}} & \small{\texttt{\href{mailto:ayan.chakraborty@epfl.ch}{ayan.chakraborty@epfl.ch}}} & \small{\texttt{\href{mailto:nicholas.sperrygrandhomme@epfl.ch}{\makecell{nicholas.sperry\\grandhomme@epfl.ch}}}} \\
    & & \\
    \textbf{Babak Falsafi} & \textbf{Martin Jaggi} & \textbf{Yunho Oh}  \\
    \small{EcoCloud, EPFL} & \small{EcoCloud, EPFL} & \small{Korea University} \\
    \small{\texttt{\href{mailto:babak.falsafi@epfl.ch}{babak.falsafi@epfl.ch}}} & \small{\texttt{\href{mailto:martin.jaggi@epfl.ch}{martin.jaggi@epfl.ch}}} & \small{\texttt{\href{mailto:yunho_oh@korea.ac.kr}{yunho\_oh@korea.ac.kr}}} \\
  \end{tabularx}
}

\begin{document}

\maketitle

\begin{abstract}
The unprecedented demand for computing resources to train DNN models has led to a search for minimal numerical encoding. Recent state-of-the-art (SOTA) proposals advocate for multi-level scaled narrow bitwidth numerical formats. In this paper, we show that single-level scaling is sufficient to maintain training accuracy while maximizing arithmetic density. We identify a previously proposed single-level scaled format for 8-bit training, Hybrid Block Floating Point (HBFP), as the optimal candidate to minimize. We perform a full-scale exploration of the HBFP design space using mathematical tools to study the interplay among various parameters and identify opportunities for even smaller encodings across layers and epochs. Based on our findings, we propose \emph{Accuracy Booster}, a mixed-mantissa HBFP technique that uses 4-bit mantissas for over $99\%$ of all arithmetic operations in training and 6-bit mantissas only in the last epoch and first/last layers. We show Accuracy Booster enables increasing arithmetic density over all other SOTA formats by at least $2.3\times$ while achieving state-of-the-art accuracies in 4-bit training.
\end{abstract}

\input{1_introduction.tex}

\input{2_arithmetic_density}

\input{3_minimizing.tex}
\input{4_experimental_results.tex}
\input{5_related_work.tex}
\input{6_conclusion.tex}

\begin{ack}
We thank Amir Yazdanbakhsh, Antoine Bosselut, Bita Darvish Rouhani, Shashwat Shrivastava, and the anonymous reviewers for their feedback and support. This work was partially supported by the SNSF project “Unified Accelerators for Post-Moore Machine Learning” (200021\_212757) and a Microsoft Research PhD Fellowship.
\end{ack}

\newpage

%bibliographystyle{abbrv}

%%%%%%%%%%%%%%%%%%%%%%%%%%%%%%%%%%%%%%%%%%%%%%%%%%%%%%%%%%%%

\appendix
\input{7_appendix.tex}

%%%%%%%%%%%%%%%%%%%%%%%%%%%%%%%%%%%%%%%%%%%%%%%%%%%%%%%%%%%%

\end{document}

%% file: 1_introduction.tex
\section{Introduction}
\label{sec:introduction}

Over the past decade, improvements in Deep Neural Network (DNN) algorithms have led to unprecedented growth in model complexity and dataset sizes. This growth has led to an exponential increase in the demand for computational resources to train DNN models which surpasses the conventional growth rate (according to Moore's Law) in computational resources in the underlying hardware. Researchers and vendors have begun to search for alternate ways to improve the arithmetic density (operations per second per unit area of silicon) and operand storage in the underlying hardware platforms.
Narrow bitwidth, low-precision numerical formats~\cite{micikevicius:mixedfp16, wang:bfloat16, sun:hybridfp8, sun:ultralow4bit, fp8:micikevicius} have emerged as a promising approach to overcome these limitations allowing for both high arithmetic density 

An ideal narrow bitwidth numerical format should maximize arithmetic density and minimize the storage requirements while maintaining FP32-level accuracy during training.
Current state-of-the-art formats fail to optimize for all three constraints below $8$ bits. 
Floating-point formats such as FP16 or BFloat16~\cite{wang:bfloat16} maintain FP32-level accuracy but suffer from low arithmetic density. 
Likewise, fixed-point formats (such as INT8) do not work well for training because gradients have widely varying magnitudes. 

Recent work has explored the use of scaled numerical formats to improve arithmetic density while maintaining high accuracy. 
Element-wise formats such as FP8 manage scaling factors in software at a coarse granularity ($\approx$ 1000) and have not been show to reach FP32-level accuracy with 4-bit fixed-point elements. 
In contrast, block-wise formats such as Block Floating Point (BFP)~\cite{koster:flexpoint, das:dynamicfixed, zhang2022fast, rouhani:msfp} manage scaling factors in hardware to strike a balance between floating point and fixed point. 
BFP organizes floating-point numbers into blocks of mantissas, enabling fixed-point arithmetic by sharing a single exponent (as a scaling factor) within a block of fixed-point mantissas. 
Because the exponent management overhead in hardware gets amortized over the block size, BFP's arithmetic density reaches that of fixed point even with fine-grain blocks ($<100$ elements). 
Unfortunately, block-based formats by themselves have not been shown to maintain FP32-level accuracy for 4-bit training. 

Similarly, in recent years, a consortium of vendors under the umbrella of the Open Compute Project~\cite{ocp:ocp,bita:mx,rouhani2023microscaling} has proposed block-wise scaled numerical formats based on shared micro-exponents (MX) allowing for two levels of fine-grained scaling on a block of mantissas. While micro-exponents optimize for storage density and have been shown to reach high accuracies with $6$ bits per element, two-level exponent management in hardware fundamentally limits their arithmetic density as compared to 4-bit fixed point.

In this paper, we ask the question of whether scaled numerical formats can reach 4-bit fixed-point arithmetic density. Without loss of generality, we use Hybrid Block Floating Point (HBFP)~\cite{drumond:hbfp}, a mixed-precision recipe using BFP for dot products and higher precision (FP32/BFloat16) for the remaining operations to achieve FP32-level accuracy. We perform a full-scale empirical parameter exploration of HBFP with mathematical tools to derive insights about the impact of BFP's parameter space on accuracy across layers and epochs. We show for the first time that with a fixed mantissa size across all epochs and layers, the minimum mantissa requirement for HBFP is 6 bits. 

We then present \emph{Accuracy Booster}, a novel mixed-mantissa training recipe that uses 4-bit mantissas in over 99\% of training operations, and 6-bit mantissas in just the last epoch and first and last layers of DNN models. We use analytic models to estimate the silicon area of arithmetic units and show that Accuracy Booster increases the arithmetic density of a training accelerator by up to $25\times$ over FP32, $5\times$ over BFloat$16$, $4\times$ over FP8, $2.3\times$ over MXFP$6$, and $2.8\times$ over MX9 while achieving state-of-the-art accuracies. Moreover, Accuracy Booster reduces the per-element storage requirement by $33\%$ from $6$ bits to $4$ bits for all but the last epoch.

%% file: 2_arithmetic_density.tex
\section{Arithmetic \& storage density}
\label{sec:density}

In this section, we present a model to compare and contrast the arithmetic and storage density of a systolic-array-based training accelerator for state-of-the-art scaled numerical formats. Our metrics of interest are arithmetic density in operations per second per unit area of silicon, and operand storage in average number of bits required to store an element. We model a canonical systolic array rather than just multiply-accumulate units to estimate the area more precisely. The area estimate is more precise because, besides the basic logic to perform training operations, block-wise scaled numerical formats also require conversion units to and back from higher precision (e.g., FP32/BFloat16) to maintain FP32-level accuracy~\cite{drumond:hbfp, bita:mx, rouhani2023microscaling}.

\input{main/figs/hardware}

Figure~\ref{fig:sys_array} depicts the anatomy of a systolic array training accelerator of $N\times N$ processing elements (PEs). Each PE computes the dot product of a group of $W$ mantissas per cycle for each HBFP block, and includes an adder to compute the resulting exponent. The systolic array can compute the product of two tiles of size $(N*W)\times N$. We choose $N=8$ for all our analyses to ensure that the number of arithmetic units ($=N^2*W)$ does not exceed typical values found in modern accelerators. The array includes F2B conversion units along the edge and B2F conversion units integrated with each PE's accumulator logic needed by block-based scaled numerical formats. Assuming all number formats take the same number of clock cycles for matrix multiplication, arithmetic density can be expressed as $(N^2*W)/(\text{Systolic Array Area})$.

Much like prior work~\cite{baalen:fp8vsint8}, we estimate silicon area as the total number of basic logic gates present in each systolic array. We assume the same clock rate for all configurations so that the number of operations per unit area becomes a proxy for arithmetic density. This approach~\cite{buyuksahin:gatecount2002} allows us to quickly iterate over various systolic array configurations without manually designing and synthesizing hardware for each numerical format. When evaluating element-wise formats, we estimate the number of logic gates for a dot product of tile $W$ elements, equal to the block size in HBFP (or MX) or first scaling level for MXFP. The element-wise formats also include exponent management logic and an accumulator per element.

\input{main/tables/num_formats}

In the rest of this paper, we refer to an HBFP configuration with $m$ mantissa bits as HBFP$m$. We corroborate prior work~\cite{wang:bfloat16, bita:mx} indicating that $8$-bit exponents are sufficient for maintaining high accuracy for block-wise numerical formats. Because the exponent's silicon footprint is negligible relative to the block sizes we consider in this study, we use $8$-bit exponents in the first scaling level in all block-wise formats without loss of generality. 

Figure~\ref{fig:fp32_hbfp} compares the arithmetic density of HBFP with varying number of mantissa bits, and state-of-the-art element-wise formats, BFloat16 and FP8, normalized to that of FP32. The figure indicates HBFP's superior arithmetic density over element-wise and coarse-grain scaled numerical formats thanks to its fixed-point mantissas. Moreover, the exponent management overhead quickly becomes negligible, with over 100 elements per block, allowing for the systolic array to reach fixed-point density. While the PE logic complexity is quadratic in the number of mantissa bits~\cite{drumond:hbfp}, we only observe a slightly superlinear improvement in asymptotic density between HBFP8 and HBFP4 because the conversion units do not scale commensurately. 

The figure also indicates that with a block size of $64$, an accelerator achieves $\geq 90\%$ of asymptotic arithmetic density improvement. Sections~\ref{sec:booster} \&~\ref{sec:experimental_results} use mathematical analysis to show that a block size of $64$ allows our numerical formats to achieve state-of-the-art accuracies. The figure illustrates that HBFP$4$ with a block size of $64$ achieves $25\times, 5\times, 4\times$ higher density than FP32, BFloat16, and FP8, respectively. Not surprisingly, these floating-point formats are fundamentally limited by the inefficiencies of floating-point multipliers and accumulators. Unlike HBFP, MX and MXFP formats opt for smaller mantissas to allow for second-level scaling factors, and as such need smaller block sizes (e.g., $16$-$32$) than HBFP to maintain high accuracy.

Table~\ref{tab:formats} compares the parameter space, the arithmetic density normalized to a baseline HBFP8, and per-element storage requirement for various state-of-the-art block-wise numerical formats. The results show that among the formats with $\approx4$ bits per element, HBFP achieves the best arithmetic density. MX formats suffer a slight degradation in density due to their two-level scaling. Nevertheless, MX formats are only used for inference because they can not reach sufficient accuracy in training~\cite{bita:mx}. In contrast, MXFP formats result in much lower arithmetic density due to two-level exponent management and a second-level scaling with a block size of one. Finally, even though MXFP$4$ achieves a slightly better arithmetic density than MXFP$6$, it has only been used in practice for weights and still requires 6-bit arithmetic~\cite{rouhani2023microscaling}.

%% file: main/figs/hardware.tex
\begin{figure}[!t]
\centering
\begin{minipage}{.52\textwidth}
  \centering
  \includegraphics[width=0.99\linewidth]{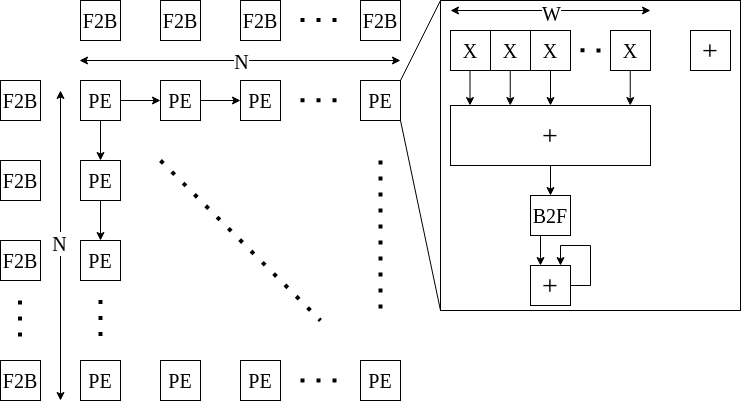}
  \caption{An HBFP systolic array accelerator.}
  \label{fig:sys_array}
\end{minipage}%
\hspace{0.04\textwidth}%
\begin{minipage}{.43\textwidth}
  \centering
  \includegraphics[height=0.72\linewidth]{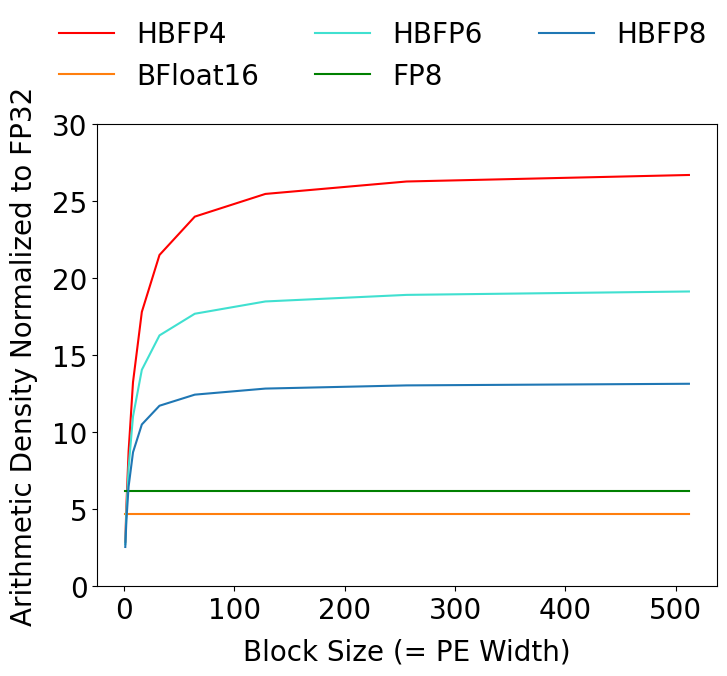}
  \caption{Normalized arithmetic density.}
  \label{fig:fp32_hbfp}
\end{minipage}
\end{figure}

%% file: main/tables/num_formats.tex
% \usepackage{multirow}
\begin{table}[ht]
\label{tab:formats}
\caption{SOTA block-wise numerical formats, their arithmetic density and storage requirements.}
\centering
  \resizebox{0.97\columnwidth}{!}{
\begin{tabular}{llccccccccc}
    \toprule
     & & HBFP8 & HBFP6 & HBFP4 & MX9  & MX6\textsuperscript{1}  & MX4\textsuperscript{1}  & MXFP8 & MXFP6 & MXFP4 \\

    \midrule
     {\multirow{2}{*}{Scaling Level 1}} & Block size    & 64    & 64    & 64    & 16   & 16   &   16   &   32    &   32    & 32    \\ \cmidrule{2-11}
      & Exponent bits & 8     & 8     & 8     & 8    & 8    & 8    & 8     & 8     & 8     \\ \midrule
     {\multirow{2}{*}{Scaling Level 2}} & Block size    & N/A   & N/A   & N/A   & 2    & 2    & 2    & 1     & 1     & 1     \\ \cmidrule{2-11}
      & Exponent bits & N/A   & N/A   & N/A   & 1    & 1    & 1    & 4     & 3     & 2     \\ \midrule 
    
    \multicolumn{2}{l}{Mantissa bits}   & 7     & 5     & 3     & 7    & 4    & 2    & 3     & 3     & 1     \\ 
      \midrule\midrule 

    \multicolumn{2}{l}{Normalized arithmetic density}   & 1.0   & 1.5   & 2.3   & 0.8  & 1.4  & 1.9  & 0.9   & 1.0   & 1.1   \\ \midrule 
    
    \multicolumn{2}{l}{Average number of bits/element}   & 8.1  & 6.1  & 4.1  & 9.0 & 6.0 & 4.0 & 8.2  & 6.2  & 4.2  \\ \midrule
   
     % {\multirow{2}{*}{Arithmetic Density}} & 16-bit adders & 1.0   & 1.4   & 2.0   & 0.8  & 1.3  & 1.6  & 0.4   & 0.5   & 0.5   \\ 
   
       \multicolumn{11}{l}{\textsuperscript{1} not used for training.} \\
     
\end{tabular}
}

\end{table}

%% file: 3_minimizing.tex
\section{Minimizing HBFP}
\label{sec:minimizing}

In this section, we focus on minimizing the HBFP encoding while achieving state-of-the-art accuracies. Unlike arithmetic density and storage requirements, which depend only on the HBFP parameter space, accuracy is also a function of the DNN model and the training algorithm. We elaborate on each of these factors impacting accuracy and use mathematical tools to derive insights about their impact. Finally, we explain our proposed technique for mixed-mantissa training to achieve 4-bit fixed-point arithmetic density.

\subsection{What factors affect accuracy?}

The primary factors impacting accuracy are the HBFP parameters including the exponent bitwidth, the mantissa bitwidth and the block size. We corroborate prior results (see Section~\ref{sec:density}) that there is no sensitivity to exponent bitwidth beyond $8$ bits as we vary the block size and the mantissa bitdwidth. As such, we keep the exponent bitdwidth fixed at $8$ and explore the space to minimize HBFP including techniques that exploit mixed-mantissa training.

The interplay between mantissa bitwidth and block size and its impact on accuracy has been reported empirically in many prior studies~\cite{rouhani:msfp,rouhani2023microscaling,drumond:hbfp}.To the best of our knowledge, this paper is the first to evaluate meticulously this interplay and use rigorous mathematical tools to derive insights about how to minimize block-wise numerical formats for training state-of-the-art models.

The interplay between mantissa bitdwidth and block size is driven by the quantization process of elements into a block of mantissa sharing an exponent. The process precludes normalization because the latter requires shifting the block of mantissas and adjusting the exponent which would negatively impact the arithmetic density (i.e., the elements are calculated as $2^{exponent}\times0.mantissa$ rather than $2^{exponent}\times1.mantissa$). As such, the precision of elements within a block is highly dependent on the largest element's exponent.

The largest exponent in a block and the mantissa bitwdith dictate the magnitude of the  interval defined by two representable consecutive numbers. This magnitude indicates that as the block size increases, the likelihood of disparity in the magnitude of elements also increases, leading to a precision loss for smaller elements in the block. Similarly, as the mantissa bitwidth decreases, the model's sensitivity to the block size increases with a corresponding increase in the interval, leading to a higher quantization error. A larger mantissa bitwidth improves resilience to quantization errors due to larger block sizes, as each element can be represented more precisely. 

Besides the HBFP parameters, other key factors impacting accuracy are the DNN model's layers and the training epochs. The HBFP parameters can be tailored across layers and epochs with mixed-precision training. As a block-wise numerical format, HBFP naturally lends itself to mixed-mantissa training, keeping a fixed exponent for all blocks, and varying the mantissa bitwidth with little impact on the accelerator datapath (e.g., with bitwidths that are multiples of two). Prior work shows that keeping the first convolution layer and the last fully connected layer in high precision is crucial for CNN models~\cite{choi:pact, zhou:dorefanet, mellempudi:mixedfp8, wang:training}. 
Similarly, in Transformers, the first layer maps input tokens to dense word vectors in the embedding layer. The last layer in DNN models generates a probability distribution for the possible outcomes of the task. These layers require a boost in precision to maintain high accuracy, allowing for reduced precision in other layers.

Likewise, prior work~\cite{rahaman:spectral, xu:frequency, fu:fractrain, fu:cpt} shows that DNNs learn low-frequency, simple components in initial epochs and high-frequency, complex components in final epochs. In light of these findings, we hypothesize that high-frequency components are more sensitive to quantization errors, requiring larger mantissa bitwidths during the final training epochs. Thus, starting with small mantissa bitwidths for generalization and switching to larger ones in final epochs for optimization can boost accuracy.

\subsection{Wasserstein distance \& loss landscapes}
\label{sec:booster}

To analyze the relationship between model accuracy, HBFP parameters, the DNN model and the training algorithm, we use two mathematical methods, Wasserstein distance and loss landscapes.
These techniques provide insights into HBFP's optimization and generalization capabilities.
The following subsections elaborate on our analysis and findings using these methods.

Preserving accuracy with narrow bitwidths depends on the numerical format's representational power. If the format does not distort tensor distributions during training, accuracy is maintained.
To study the distortion in distributions, we measure the similarity between BFP and FP32 tensors from various layers using Wasserstein distance, which is defined as:
\begin{equation}
    W(P,Q)=\inf_{\gamma \in \Pi(P,Q)} \mathbb{E}_{(x,y) \sim \gamma}[||x-y||],
    \label{eq:earthmovers}
\end{equation}
where $\Pi(P,Q)$ is the set of all joint distributions $\gamma(x,y)$ whose marginal distributions are equal to $P$ and $Q$. 
$\gamma(x, y)$ represents the amount of mass that must be transported from $x$ to $y$ to transform $P$ to $Q$~\cite{arjovsky:earthmover}. 
Unlike KL Divergence, which is commonly used to compare quantized tensors to their full-precision counterparts~\cite{rouhani:msfp, nvidia:tensorrt}, Wasserstein distance is symmetric, and thus a valid mathematical metric.
Additionally, DNNs frequently deal with distributions where KL Divergence is not defined or infinite, necessitating the addition of a noise term that can cause disturbances.
Our findings show that R-squared values between accuracy and various Wasserstein distances are approximately 0.99, validating the robustness of Wasserstein distance.

Figure~\ref{fig:earthmover} compares Wasserstein distance various HBFP configurations to that of FP32 for four layers of ResNet20 trained on CIFAR10. These layers include the first (conv1.weight), the last (fc.weight), and two representative convolutional layers in the middle of the model. We report power-of-two block sizes that are square numbers (i.e., $16$, $64$, $256$) as blocks are formed as tiles. We also evaluate a block size of $576$ as the baseline from~\cite{drumond:hbfp}. We observe that HBFP$4$ has approximately a $3.5\times$ higher distance than HBFP$6$, indicating more distorted tensor distributions.
As a smaller Wasserstein distance indicates a closer match to FP32 accuracy, we hypothesize that HBFP$6$ is more likely to achieve accuracies comparable to FP32 than HBFP$4$.

\input{main/figs/tools}

The figure also shows that HBFP$6$ maintains consistent Wasserstein distances across various block sizes, while HBFP$4$ distances increase rapidly with block size. 
This observation indicates that HBFP$6$ accuracy will be invariant across various block sizes, whereas HBFP$4$ accuracy will degrade with larger block sizes. Based on these observations, we hypothesize HBFP$6$ is the smallest standalone HBFP format that will achieve competitive accuracies with FP32, with minimum sensitivity to block size.

Figure~\ref{fig:earthmover} also provides important insights into the behavior of the first and last layers of a CNN model. These layers show higher Wasserstein distances, indicating they are less resilient to quantization and should be kept at higher precision. 
Additionally, the increase in Wasserstein distance for these layers is significantly greater for HBFP$4$ compared to HBFP$6$, suggesting that these layers are more sensitive to HBFP$4$ and should be maintained at a precision of at least HBFP$6$.

To gain deeper insights into the interplay between generalization and optimization power of HBFP, we analyze log-scale loss landscapes for various configurations~\cite{li:losslandscape}. 
Visualizing loss landscapes highlights two key features: (1) the lower the curve's minimum, the higher the accuracy of that configuration, and (2) the flatter the curve's minimum (assuming it is sufficiently low) the better its ability to generalize. Ideally, we seek an HBFP configuration that matches FP32's minimum and has a flatter curve. 

The landscapes are generated across two axes, each representing a random direction. These two random directions are orthogonal due to the high-dimensional vector space, and they form a grid. For each training run, the loss value corresponding to the minimizer's current state is positioned at the grid's center $(0,0)$.
As the loss landscape shows similar behavior in both directions, we plot the landscape sliced along a single axis for clarity in Figure~\ref{fig:landscapes} (3D versions are available in Appendix~\ref{app:loss_landscapes}).

The figure indicates that the minimum of the HBFP$6$ curve is fairly close to that of FP32. In contrast, the HBFP$4$ curve's minimum is significantly higher than that of FP32 even though its relatively flat shape may suggest better generalization. These findings are consistent with the Wasserstein distance analysis above.

Finally, to test the effect of keeping the first and last layers in higher precision, we use HBFP$6$ for these layers and HBFP$4$ for the remaining layers. This configuration, referred to as "HBFP$4$+Layers" in Figure~\ref{fig:landscapes}, shows a sharper and lower curve compared to HBFP$4$.  However, the model's generalization and optimization powers are still imbalanced, resulting in convergence to another suboptimal local minimum.

To bridge the gap between generalization and optimization powers, we leverage the insight that the final epochs require higher precision. We introduce Accuracy Booster, a mixed-mantissa HBFP technique that uses HBFP$6$ only in the last epoch and first/last layers, and HBFP$4$ for the rest. We hypothesize that employing HBFP$6$ for the last epoch is sufficient to enhance the model accuracy, while the preceding epochs use HBFP$4$ to help the model generalize and reach a sufficiently low loss value. Switching to HBFP$6$ for the final epoch optimizes and fine-tunes the model, boosting accuracy close to FP32 levels.

The loss landscape for Accuracy Booster (referred to as "Booster" in Figure~\ref{fig:landscapes} for brevity) supports our hypothesis.
We observe that the Booster curve approaches the HBFP$6$ and FP32 curves, and the flatness at the minima due to initial HBFP4 training indicates higher generalization power than FP32. This demonstrates that Accuracy Booster effectively uses minimal HBFP encoding and is the recipe to reach the sweet spot between generalization and optimization.

%% file: main/figs/tools.tex
\begin{figure}[!t]
\centering
\begin{minipage}{.48\textwidth}
  \centering
\includegraphics[width=0.8\linewidth,height=.67\linewidth]{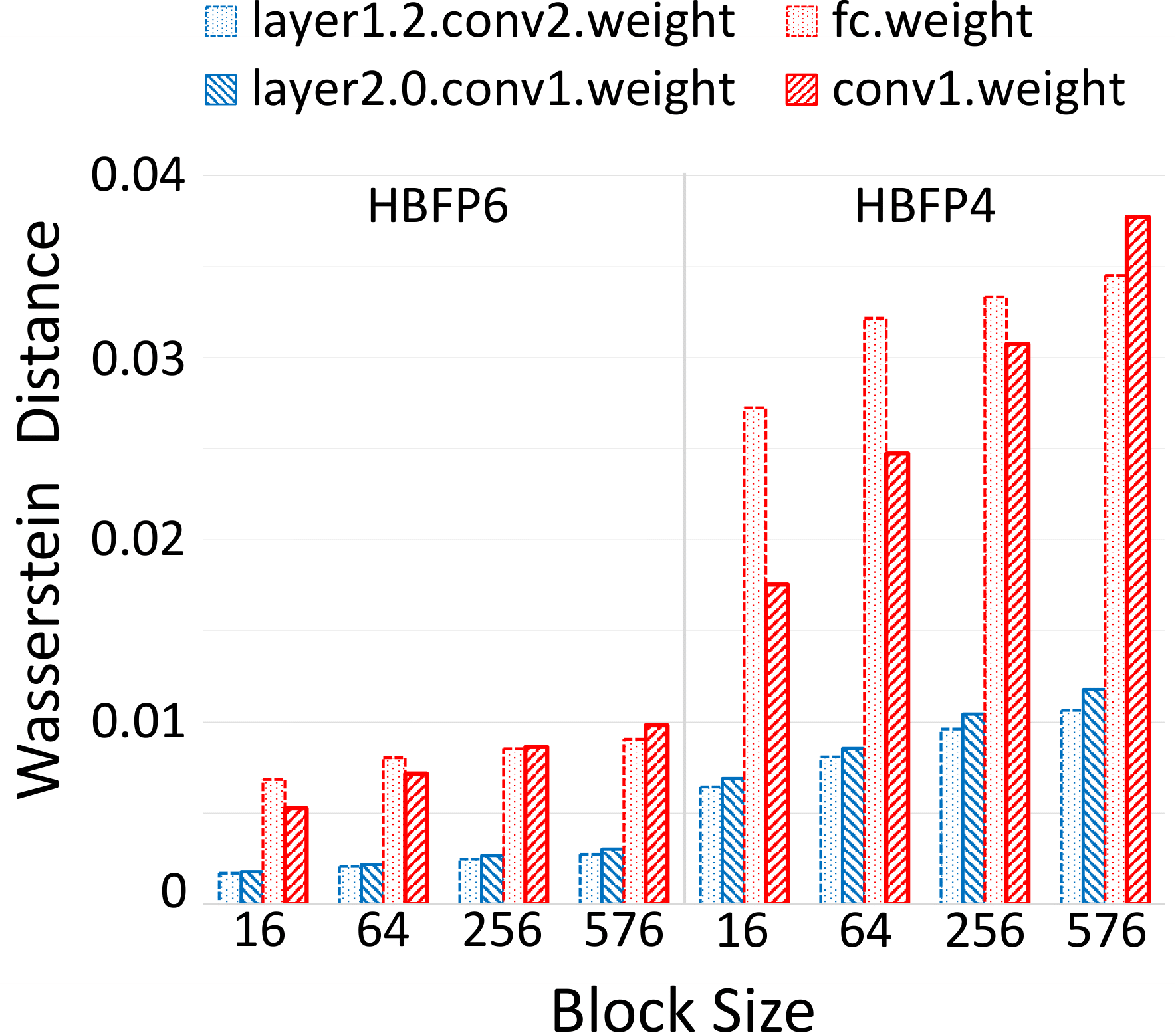}
  \caption{Wasserstein distance relative to FP32.}
  \label{fig:earthmover}
\end{minipage}%
\hspace{0.04\textwidth}%
\begin{minipage}{.48\textwidth}
  \centering
  \includegraphics[height=.67\linewidth]{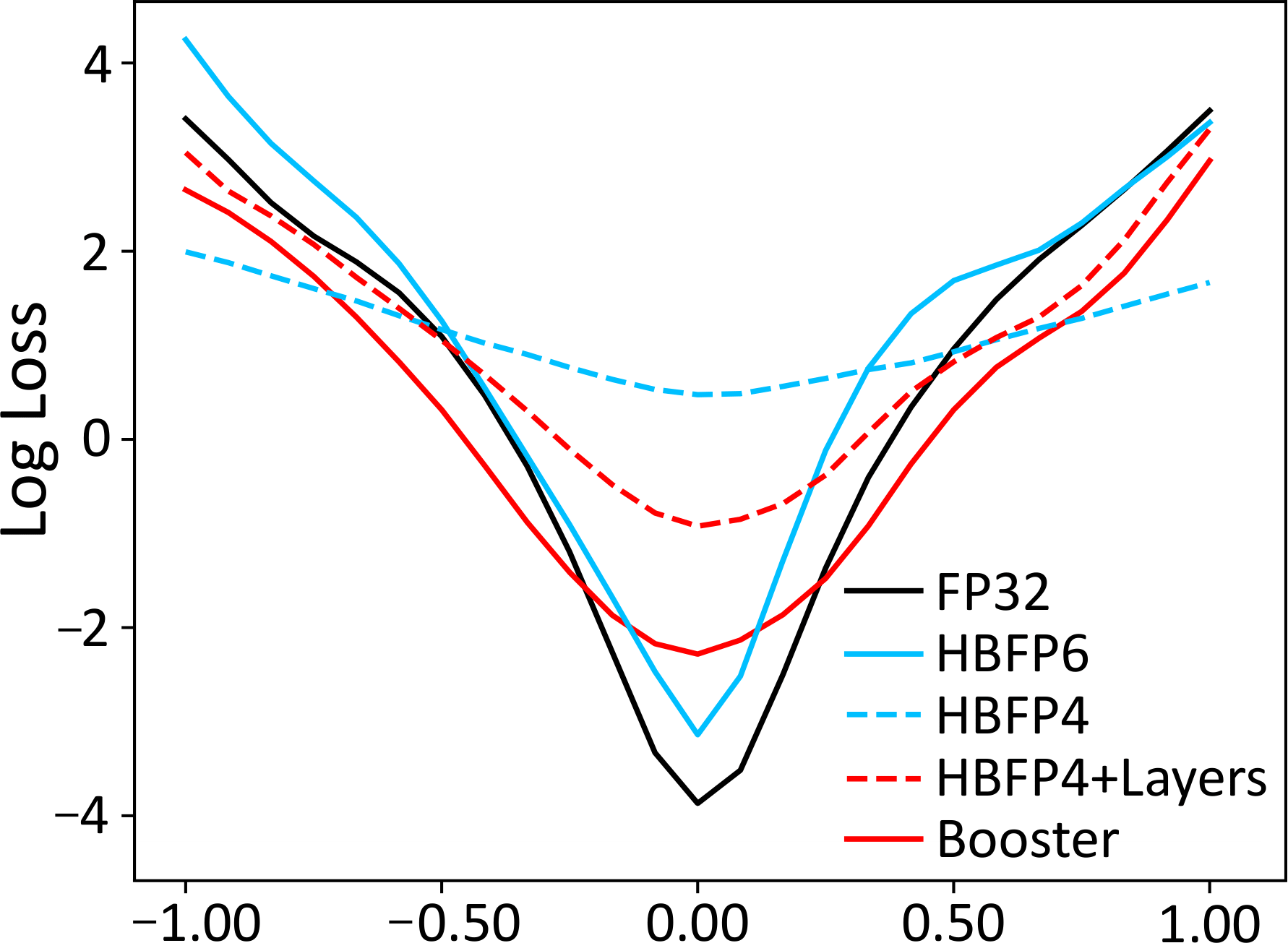}
  \caption{Loss landscapes for ResNet20.}
  \label{fig:landscapes}
\end{minipage}
\end{figure}

%% file: 4_experimental_results.tex
\section{Experimental results}
\label{sec:experimental_results}

\input{main/tables/accuracies}

Our analysis in Section~\ref{sec:minimizing} suggest the following two hypotheses: (1) HBFP$6$ is the minimum single-mantissa encoding to achieve FP32-level accuracies in training, and (2) Accuracy Booster allows for 4-bit training for all but the last epoch while achieving FP32-level accuracies. 

We test these hypotheses on state-of-the-art models and datasets for vision and language tasks. We train ResNet20/50/74~\cite{he:resnet}, and DenseNet40~\cite{huang:densenet} on CIFAR10 and CIFAR100~\cite{krizhevsky:cifar}, and ResNet50 on ImageNet~\cite{deng2009imagenet} for image classification. 
For Transformers, we train DeiT-Tiny~\cite{touvron2021training} on CIFAR-10, Transformer-Base~\cite{vaswani:attention} on IWSLT’14 German-English~\cite{iwslt_dataset}, BERT-Base~\cite{devlin:bert} on English Wikipedia~\cite{wikidump} and BookCorpus~\cite{zhu:bookcorpus} for masked language modeling, and GPT-2-124M~\cite{nanogpt} on OpenWebText~\cite{openwebtext}.

We report all the hyperparameters we use for our experiments in Appendix~\ref{app:hyperparams}. Moreover, we tune the FP32 models' hyperparameters and use the same hyperparameters for HBFP training, showing that our technique can be used out of the box without further tuning. 

\subsection{Baseline Single-Mantissa HBFP}

Table~\ref{tab:accuracies} shows the Top1 validation accuracies for the CNN models with varying HBFP parameters and their arithmetic density normalized to FP32. The table corroborates our hypothesis that HBFP$6$ is the minimal single-mantissa HBFP configuration achieving accuracies within $2\%$ of FP32 across all models and datasets for block sizes up to $256$. Larger block sizes (e.g., $576$ reported in prior work~\cite{drumond:hbfp} reach diminishing returns in arithmetic density, and more importantly negatively impact accuracy for ResNet20 and ResNet50. In Section~\ref{subsec:booster_results}, we also present results for Transformer-based models with HBFP$6$ achieving FP32-level accuracy.

In contrast to HBFP$6$, HBFP$4$ offers higher arithmetic density but falls short of reaching FP32 accuracies for any block size.\footnote{We report HBFP$5$ in Appendix~\ref{app:hbfp5accuracies} to corroborate our conclusion about HBFP$6$ being the baseline single-mantissa encoding with FP32-level accuracies.}. As the accuracy loss for ResNet50 and ResNet74 on CIFAR100 is considerably high, we did not train these models with HBFP$4$. Even for small models like ResNet20, the accuracy drops more than $9\%$ with a block size of $16$.
We observe that for HBFP$4$, the sensitivity to block size increases for all the models because of higher distortion levels in tensor distributions.

\subsection{Accuracy Booster}
\label{subsec:booster_results}

\input{main/tables/booster}

\input{main/tables/booster2}

In this section, we evaluate Accuracy Booster on various CNN and Transformer-based models, presented in Table~\ref{tab:booster} and Table~\ref{tab:booster2}, respectively. 
We choose a block size of $64$ as the sweet spot for Accuracy Booster, considering the HBFP hardware model.
A block size of $64$ is within $90\%$ of the maximum arithmetic density gain while achieving accuracies with less than $1\%$ degradation for standalone HBFP$6$. We use HBFP$6$ in the last epoch of training and HBFP$4$ for the rest of the epochs in all CNNs, DeiT-Tiny, and Transformer-Base. For the language modeling experiments, we use HBFP$6$ in the last $5$k iterations for BERT-Base and the last $2$k iterations for GPT-2.
As an ablation study, we also used HBFP$6$ in the last 10 epochs for CNNs to test for further accuracy improvements.

\begin{figure*}[!t]
\centering
\begin{minipage}{.48\textwidth}
  \centering
  \includegraphics[width=.85\linewidth,height=.7\linewidth]{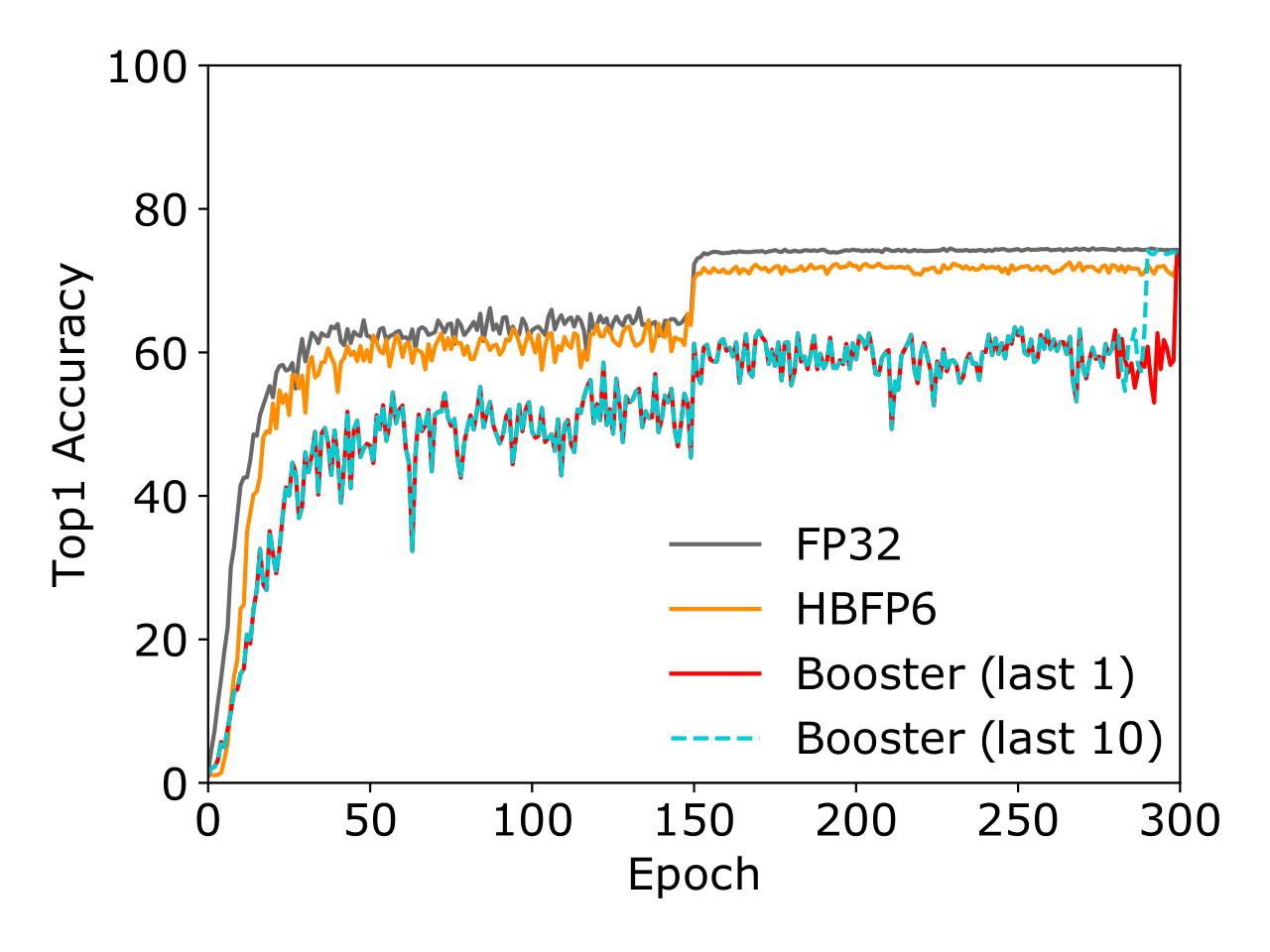}
  % \caption{Top1 Accuracies for ResNet74 on \\ CIFAR100 with various training techniques.}
  % \label{fig:cnn_boosterplots}
\end{minipage}%
\hspace{0.04\textwidth}%
\begin{minipage}{.48\textwidth}
  \centering
  \includegraphics[width=.85\linewidth,height=.7\linewidth]{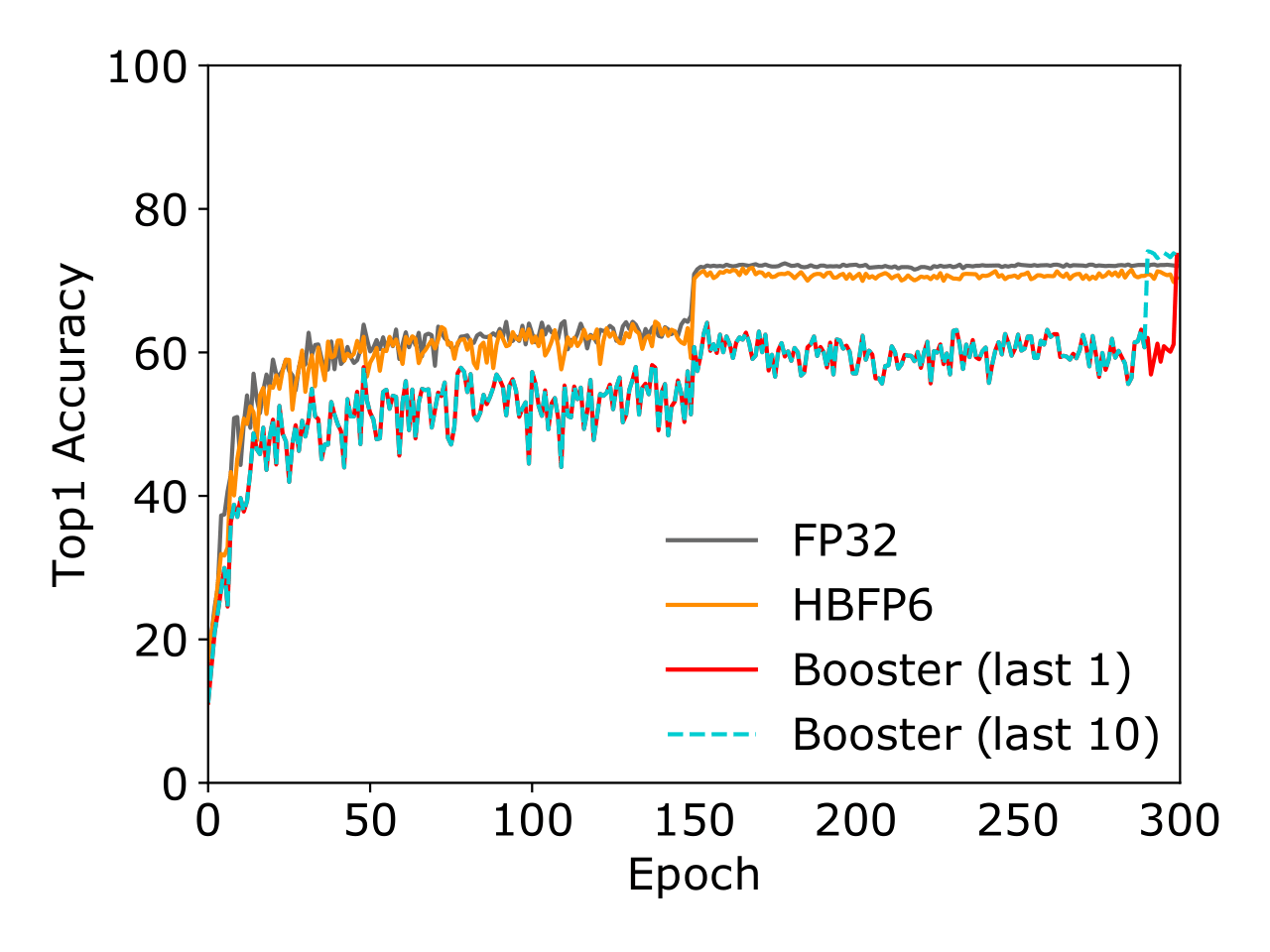}
  % \caption{Top1 Accuracies for DenseNet40 on CIFAR100 with various training techniques.}
  % \label{fig:cnn_boosterplots}
\end{minipage}
\caption{Top1 Accuracy training curves for ResNet74 and DenseNet40 on CIFAR100}
\label{fig:cnn_boosterplots2}
\end{figure*}

We observe that for all CNN models, Accuracy Booster performs similarly to FP32, showing a huge jump in accuracies compared to standalone HBFP$4$ training. Similarly, for BERT-Base and Transformer-Base, Accuracy Booster outperforms FP32.  For GPT-2 and DeiT-Tiny, Accuracy Booster incurs a minor accuracy loss, but it is still within the $1\%$ error margin of FP32 accuracies. When we keep the last $10$ epochs in HBFP$6$, we observe that the accuracies slightly increase, which may not be worth the extra computation depending on application needs.

Figure~\ref{fig:cnn_boosterplots2} shows the validation accuracy curves of ResNet74 and DenseNet40 on CIFAR100 with FP32, HBFP$6$, and Accuracy Booster (the curves for the other models are in Appendix~\ref{app:curves}). 
For the Booster curves, we observe a noticeable sharp increase in accuracy towards the end for all models, demonstrating the significance of increasing precision in the final epoch.
Overall, the training curves for Accuracy Boosters follow similar trends across all epochs compared to FP32 and HBFP6, albeit a bit noisier due to the reduced precision.

Our method achieves only $2\%$ accuracy degradation with ResNet50 on ImageNet by using only 4-bit fixed-point arithmetic.
Sun et al.~\cite{sun:ultralow4bit} propose 4-bit training by using INT4xINT4 arithmetic for forward pass and INT4xFP4 arithmetic for backward pass. 
However, their design requires support for both multiplication units, thus incurs a higher silicon area. 
Similarly, a recent paper by Chmiel et al.~\cite{chmiel:4bitint} applies the same regime but propose a more efficient way for INT4xFP4 multiplication and the resulting MAC unit is equivalent to FP7.
These two papers report $2.3\%$ and $1.1\%$ degradation on ImageNet respectively with roughly $2.5\times$ less arithmetic density compared to our method.

%% file: main/tables/accuracies.tex
% \captionsetup{width=0.7\textwidth}
% \captionsetup{justification=centering}
\begin{table*}[b]
  \caption{Top-1 validation accuracies of various CNN models for various HBFP configurations.}
  \small
  \label{tab:accuracies}
  \centering
  \resizebox{0.97\columnwidth}{!}{
  \begin{tabular}{ccccccccc}
    \toprule

     & & &
    \multicolumn{2}{c}{CIFAR10} & \multicolumn{4}{c}{CIFAR100}                   \\
    
    \cmidrule(l{4pt}r{4pt}){4-5}
    \cmidrule(l{4pt}r{4pt}){6-9} 
    \makecell{Number\\Format} & Block Size & \makecell{Normalized \\Arithmetic Density}  & ResNet20 & ResNet74 & ResNet50 & ResNet74 & DenseNet40 \\
    \cmidrule(l{3pt}r{3pt}){1-1} 
    \cmidrule(l{3pt}r{3pt}){2-2} 
    \cmidrule(l{3pt}r{3pt}){3-3} 
    \cmidrule(l{3pt}r{3pt}){4-4} 
    \cmidrule(l{3pt}r{3pt}){5-5} 
    \cmidrule(l{3pt}r{3pt}){6-6} 
    \cmidrule(l{3pt}r{3pt}){7-7} 
    \cmidrule(l{3pt}r{3pt}){8-8} 
    \cmidrule(l{3pt}r{3pt}){9-9}

    FP32 & - & $1$ & $91.7$ & $93.6$ & $74.1$ & $74.6$ & $72.4$     \\
    \midrule
    HBFP8 & {576} & $13$ & $91.5$ & $93.4$ & $73.8$ & $74.3$ & $73.7$     \\
    \midrule
    
    \multirow{4}{*}{HBFP$6$} 
     & {$16$} & $15$ & $91.1$ & $93.4$ & $73.1$ & $73.5$ & $72.1$     \\
    % & {$25$} / $17$ & $91.1$ & $92.5$ & $72.6$ & $73.2$ & $71.8$    \\
    % & {$36$} / $17$ & $91.3$ & $92.6$ & $72.5$ & $72.9$ & $71.8$    \\
    % & {$49$} / $18$ & $91.3$ & $92.9$ & $72.3$ & $72.4$ & $71.9$    \\
     & {$64$} & $18$ & $91.1$ & $92.9$ & $72.1$ & $72.4$ & $71.8$    \\
     & {$256$} & $19$ & $91.4$ & $92.8$ & $72.0$ & $72.5$ & $71.5$    \\
     & {$576$} & $20$ & $90.7$ & $92.2$ & $70.8$ & $72.5$ & $71.1$   \\
    \midrule

    % \multirow{7}{*}{HBFP$5$} 
    %  & {$16$} / $17.3$ & $90.2$ & $92.2$ & $67.2$ & $70.5$ & $69.4$    \\
    %  & {$25$} / $19.2$ & $90.2$ & $92.0$ & $67.9$ & $69.0$ & $70.0$     \\
    %  & {$36$} / $20.3$ & $90.1$ & $92.0$ & $67.1$ & $69.9$ & $69.0$     \\
    %  & {$49$} / $21.1$ & $89.7$ & $92.0$ & $68.7$ & $68.7$ & $69.3$     \\
    %  & {$64$} / $21.7$ & $89.7$ & $91.3$ & $66.8$ & $69.3$ & $68.9$     \\
    %  & {$256$} / $23.1$ & $89.1$ & $91.1$ & $63.0$ & $66.6$ & $67.8$    \\
    %  & {$576$} / $23.4$ & $88.9$ & $91.0$ & $64.6$ & $64.4$ & $67.6$     \\
    % \midrule
    
    % \multirow{7}{*}{\makecell{HBFP$5$ with \\ first and last\\layers in FP32}} 
    %  & {16} / $-$ & $90.80$ & $92.56$ & $71.23$ & $72.31$ & $70.90$     \\
    %  & {25} / $-$ & $90.62$ & $91.79$ & $71.76$ & $72.58$ & $70.52$     \\
    %  & {36} / $-$ & $90.62$ & $92.73$ & $71.45$ & $72.26$ & $70.01$     \\
    %  & {49} / $-$ & $90.49$ & $92.02$ & $70.51$ & $72.07$ & $70.16$     \\
    %  & {64} / $-$ & $90.31$ & $92.41$ & $69.59$ & $72.09$ & $70.10$    \\
    %  & {256} / $-$ & $90.07$ & $92.11$ & $68.76$ & $70.20$ & $69.21$    \\
    %  & {576} / $-$ & $90.37$ & $91.06$ & $68.66$ & $69.14$ & $68.78$     \\
    % \midrule
        
    \multirow{4}{*}{HBFP$4$} 
     & {$16$} & $20$ & $82.6$ & $76.9$ & - & - & $63.7$    \\
   %  & {$25$} / $22$ & $81.8$ & $78.6$ & - & - & $64.3$     \\
   %  & {$36$} / $23$ & $80.8$ & $76.6$ & - & - & $63.3$     \\
   %  & {$49$} / $24$ & $79.3$ & $71.2$ & - & - & $65.6$     \\
     & {$64$} & $25$ & $80.2$ & $74.4$ & - & - & $62.4$    \\
     & {$256$} & $27$ & $77.0$ & $60.7$ & - & - & $60.0$   \\
     & {$576$} & $27$ & $75.3$ & $66.7$ & - & - & $59.8$     \\
    \midrule\midrule

    \multicolumn{3}{c}{{\makecell{Total Number of FLOPs \\ required to train the model}}} & $41$M & $174$M & $119$M & $326$M & $542$M \\   
    %\multicolumn{3}{c}{{Number of Model Parameters}} & $270$K & $38$M & $25$M & $38$M & $1$M     \\
   
    \bottomrule
  \end{tabular}
  }
\end{table*}

%% file: main/tables/booster.tex
% \captionsetup{width=0.7\textwidth}
% \captionsetup{justification=centering}
\begin{table}[b]
  \caption{Top-1 validation accuracies of various CNN models for Accuracy Booster}
  \label{tab:booster}
  \centering
  \resizebox{0.72\columnwidth}{!}{
  \begin{tabular}{ccccccc}
    \toprule
    &  & 
    \multicolumn{2}{c}{CIFAR10}  & \multicolumn{1}{c}{ImageNet}    & \multicolumn{2}{c}{CIFAR100}                   \\
    
    \cmidrule(l{4pt}r{4pt}){3-4}
    \cmidrule(l{4pt}r{4pt}){5-5} 
    \cmidrule(l{4pt}r{4pt}){6-7} 
    \multicolumn{2}{c}{ } & ResNet20 & ResNet74 & ResNet50 & ResNet74 & DenseNet40\\
    
    \cmidrule(l{3pt}r{3pt}){3-3} 
    \cmidrule(l{3pt}r{3pt}){4-4} 
    \cmidrule(l{3pt}r{3pt}){5-5} 
    \cmidrule(l{3pt}r{3pt}){6-6} 
    \cmidrule(l{3pt}r{3pt}){7-7} 
     
    % \multicolumn{2}{c}{Only last} & $49$ & $91.45$ & $92.85$ & \boldmath$74.25$ & $74.39$ & \boldmath$74.06$  \\
    % \multicolumn{2}{c}{Only last} & $256$ & $90.76$ & $92.40$ & $72.78$ & $74.32$ & \boldmath$74.77$  \\
    % \multicolumn{2}{c}{Last 10} & $49$ & $91.69$ & $92.87$ & \boldmath$74.50$ & \boldmath$74.45$ & \boldmath$74.26$  \\    
    \multicolumn{2}{c}{Booster (last 1)} & $91.2$ & $92.6$ & $74.2$ & $73.7$ & $73.6$ \\
    \multicolumn{2}{c}{Booster (last 10)} & $91.4$ & $93.0$ & $74.3$ & $74.3$ & $74.1$  \\
    %\multicolumn{2}{c}{First and last 10} & $576$ & $87.84$ & $71.11$ & $63.50$ & $67.25$ & \boldmath$72.79$  \\
    
    \midrule

    \multicolumn{2}{c}{{FP32}}  & $91.7$ & $93.6$ & $76.3$ & $74.6$ & $72.4$    \\
        
    %\midrule

    %\multicolumn{3}{c}{{Power consumption ratio (FP32/Booster)}} & $28.2$ & $35.0$ & $36.5$ & $38.0$ & $72.42$    \\
    \bottomrule
  \end{tabular}
  }
\end{table}

%% file: main/tables/booster2.tex
% \captionsetup{width=0.7\textwidth}
% \captionsetup{justification=centering}
\begin{table}[t]
  \caption{Evaluation of various Transformer-based models for Accuracy Booster and HBFP6. }
  \label{tab:booster2}
  \centering
  \resizebox{0.77\columnwidth}{!}{
  \begin{tabular}{cccccc}
    \toprule
    
    \multicolumn{2}{c}{ } & \makecell{BERT-Base\\(Perplexity)} & \makecell{GPT-2\\(Loss)}  & \makecell{DeiT-Tiny \\(Accuracy)}  & \makecell{Transformer-Base \\(BLEU)} \\
    
    \cmidrule(l{3pt}r{3pt}){3-3} 
    \cmidrule(l{3pt}r{3pt}){4-4} 
    \cmidrule(l{3pt}r{3pt}){5-5} 
    \cmidrule(l{3pt}r{3pt}){6-6} 
     
    % \multicolumn{2}{c}{Only last} & $49$ & $91.45$ & $92.85$ & \boldmath$74.25$ & $74.39$ & \boldmath$74.06$  \\
    % \multicolumn{2}{c}{Only last} & $256$ & $90.76$ & $92.40$ & $72.78$ & $74.32$ & \boldmath$74.77$  \\
    % \multicolumn{2}{c}{Last 10} & $49$ & $91.69$ & $92.87$ & \boldmath$74.50$ & \boldmath$74.45$ & \boldmath$74.26$  \\    
   
    \multicolumn{2}{c}{{FP32}}  & $3.3$ & $3.1$ & $93.8$  & $34.8$  \\

    \midrule
    
    \multicolumn{2}{c}{{HBFP6}}  & $3.3$ & $3.2$ & $93.7$  & $34.5$  \\
    
    \midrule
    
    \multicolumn{2}{c}{Booster} & $3.2$ & $3.2$ & $92.4$  & $36.1$ \\
    %\multicolumn{2}{c}{First and last 10} & $576$ & $87.84$ & $71.11$ & $63.50$ & $67.25$ & \boldmath$72.79$  \\
    
    \midrule\midrule

    \multicolumn{2}{c}{\makecell{Number of Model Parameters}}  & $110$M & $120$M & $5$M  & $65$M  \\
        
    %\midrule

    %\multicolumn{3}{c}{{Power consumption ratio (FP32/Booster)}} & $28.2$ & $35.0$ & $36.5$ & $38.0$ & $72.42$    \\
    \bottomrule
  \end{tabular}
  }
\end{table}

%% file: 5_related_work.tex
\section{Related work}
\label{sec:relatedwork}

In recent years, there has been a significant amount of research on inference and training~\cite{wang:bfloat16, micikevicius:mixedfp16, rastegari:xnornet, zhou:dorefanet, sun:hybridfp8, lin:fixedpoint, courbariaux:lowprec, mellempudi:mixedfp8, sun:ultralow4bit, dettmers:inference} with narrow numerical encodings.
The most recent work on FP8~\cite{fp8:micikevicius} empirically shows accurate training and inference for various models including Transformers.
However, all of these methods suffer from the intrinsically low arithmetic density of floating-point arithmetic.
Additionally, these methods usually cannot go lower than 8 bits or require careful tuning.
Recent research advocates the use of BFP for DNN training~\cite{drumond:thesis, koster:flexpoint, das:dynamicfixed, fox:minifloat} and inference~\cite{rouhani:msfp} to overcome the limitations of floating-point formats.
Drumond et al.~\cite{drumond:hbfp} combine BFP with floating point to propose HBFP.
Recently, Rouhani et al.~\cite{bita:mx} have proposed a new format based on shared micro-exponents (MX), which uses two levels of fine-grained scaling on blocks of fixed point numbers.
A follow-up work~\cite{rouhani2023microscaling} introduces MXFP6, where a block of FP6 elements share an $8$-bit exponent.
In this paper, we argue that further reducing the mantissa bitwidth in HBFP results in a significant improvement in arithmetic density for DNN training, and there is a wide design space left unexplored.

Mixed-precision training has emerged as a popular technique to recover the information loss caused by quantization.
Several techniques~\cite{khoram_adaptivequant_2018, yang:fracbits, shen:fractionalskip, choi:pact, zhou:dorefanet, mellempudi:mixedfp8, wang:training} vary precision layer-wise, using higher precision for more important layers. 
Fu et al.~\cite{fu:cpt} employ fixed-point arithmetic with varying bitwidths across epochs during training.
Some techniques~\cite{fu:fractrain, zhang2022fast, noh:flexblock} combine layer-wise and epoch-wise approaches and vary the precision adaptively per epoch and layer at the same time using various control mechanisms.
Recently proposed techniques for $4$-bit training~\cite{sun:ultralow4bit, chmiel:4bitint} utilize different arithmetic units for the forward and backward passes.
Although these aforementioned studies use leaner arithmetic for a fraction of the training process, they fail to make leaner arithmetic the common case of training, requiring additional arithmetic units.

Recent work~\cite{dettmers:8bit} suggests that the optimizer states during training can be reduced to $8$ bits by using block-wise quantization.
This observation is in line with our work, applying quantization by extracting the largest exponent per block.
Similarly, FAST~\cite{zhang2022fast} uses a BFP-based layer-wise mixed-precision approach using $2$ and $4$-bit mantissas.
Unlike our work, FAST requires fine-tuning several additional hyperparameters, making it difficult to apply to other DNN models.
Another BFP-based work, FlexBlock~\cite{noh:flexblock}, uses $4$ and $8$-bit mantissa with various block sizes but also needs higher-precision BFP formats for weight gradient calculations.

%% file: 6_conclusion.tex
\section{Discussion}
\label{sec:discussion}
\paragraph{Hardware Implications.}
Accuracy Booster primarily employs HBFP$4$ with a limited number of HBFP$6$ operations and can be implemented efficiently without requiring reconfigurable hardware. 
The $6$-bit multiplications required for HBFP$6$ can be emulated on $4$-bit HBFP processing elements in four steps (cycles). 
 This process involves separating the four lower bits and the two higher bits that constitute the $6$-bit mantissa. 
 Let $A$ and $B$ be two mantissa vectors of two HBFP$6$ blocks. 
 $A_L$ denotes the four lower bits and $A_H$ denotes the two higher bits of mantissas, such that $A=2^4 \cdot A_H + A_L$
 Multiplication between two HBFP$6$ blocks of mantissas $A\times B$ can be calculated in four cycles as follows:
 \begin{equation*}
     A \times B = 2^8 \cdot A_H \times B_H + 2^4 \cdot A_H \times B_L + 2^4 \cdot A_L \times B_H + A_L \times B_L
 \end{equation*}
Each of these four multiplications can be executed using the $4$-bit arithmetic units and accumulated by shifting units with a minimal area cost, albeit at a latency increase of $4\times$ per operation.
However, as previously discussed, these $6$-bit operations occur infrequently ($<1\%$), so the overall impact of throughput degradation due to these operations is negligible. 
Hence, we do not explicitly target reconfigurable systems; instead, we focus on fixed arithmetic units for $4$-bit operations.

\paragraph{Limitations.}
One notable constraint is the dependency on specific hardware configurations that can support mixed-mantissa arithmetic efficiently, especially for operand storage. 
While the HBFP format enhances arithmetic density, it requires hardware support to handle the separation and combination of different bitwidth mantissas seamlessly. 
This could limit the applicability of Accuracy Booster to environments where such hardware is available or feasible to implement. 
Additionally, although we report our results on an extensive set of models, we do not experiment with large language models due to their high training overhead. 
Prior work~\cite{dettmers:8bit} suggests that the outlier features in large models may hurt model accuracies of leaner formats.
There are several proposals to handle outliers in large models, and we leave this study on how to leverage these proposals to handle outliers using HBFP for future work.

\paragraph{Impact.}
By enabling 4-bit fixed-point arithmetic, Accuracy Booster dramatically increases arithmetic density, reducing the computational resources and power consumption required for training DNN models. 
This enhancement can lead to more energy-efficient training processes, making it feasible to train complex models on resource-constrained devices such as edge computing hardware and mobile devices.
Additionally, the insights gained from the mixed-mantissa approach may inspire further innovations in numerical formats and training algorithms, contributing to the ongoing advancement of efficient AI technology.

\section{Conclusion}
\label{sec:conclusion}

In this paper, we explore the extensive parameter space of BFP within the context of HBFP, delving into the interplay between block size and mantissa bitwidth. 
Leveraging mathematical tools such as Wasserstein distance and loss landscapes, we gain unique insights into the effects of training with block-wise formats on accuracy. 
Building on these insights, we introduce Accuracy Booster, a mixed-mantissa training recipe using HBFP$6$ only in the last epoch and first/last layers and HBFP$4$ for the rest. 
A significant contribution of our work lies in identifying opportunities for mixed-mantissa BFP encodings across layers and epochs during training, a previously unexplored dimension. 

Our work not only expands the understanding of BFP but also paves the way for storage- and area-efficient DNN training.
By providing a path to achieve $4$-bit training with fixed-point arithmetic, we bridge the gap between storage optimization and arithmetic density. 
Our method uses $4$ bits per element on average and increases the arithmetic density by up to $4\times$ compared to FP8 and up to $2.3\times$ compared to MXFP6 while achieving SOTA accuracies.
This research contributes to the ongoing efforts in developing high-performance and resource-efficient DNN models, providing valuable insights that could shape future hardware-software co-design strategies for DNN training.

%% file: 7_appendix.tex
\newpage
\section{HBFP5 accuracies}
\label{app:hbfp5accuracies}
We report the top-1 validation accuracies of various CNN models for HBFP5 (along with the other configurations for ease of comparison) to corroborate our conclusion about HBFP6 being the baseline single-mantissa encoding with FP32-level accuracies.
\input{main/tables/hbfp5}

\section{Hyperparameters for training}
\label{app:hyperparams}

The CNN models for image classification are trained for $160$ epochs each on CIFAR10, and $300$ epochs each on CIFAR100.
DeiT-Tiny is trained for $300$ epochs, and Transformer-Base for $70$ epochs.
We used the HuggingFace Transformers library~\cite{huggingface:transformers} for BERT-Base training and nanoGPT~\cite{karpathy:nanogpt} for GPT2 training, and we trained these models for $500$k and $12$k iterations, respectively.
We use NVIDIA V100 32GB and A100 80GB GPUs in our evaluations. 
For large models and datasets, we train on multiple GPUs by using data parallelism.
We report the total amount of computations required to train all CNN models in Table~\ref{tab:accuracies}, and the number of parameters of Transformer-based models in Table~\ref{tab:booster2}

We use FP32 precision as the baseline for model accuracies.
For the image classification experiments, we report the Top1 validation accuracies; for machine translation, we report the BLEU scores, for language modeling we report the perplexities.
Moreover, to show the impact of our method, we tune the hyperparameters of FP32 models and then train the same models from scratch with the same hyperparameters in HBFP, showing that our method can be used out of the box without further hyperparameter tuning. 

\subsection {Image classification experiments}
\label{app:hyperparams_image}
We report the hyperparameters we use for all our CNN experiments in Table~\ref{tab:cnnhyperparam}. All convolutional layer weights are initialized using  a zero-mean Gaussian distribution whose standard deviation (std) is $\sqrt{2n_l}$ ($n_l$ is the number of activations in the output), a method introduced in~\cite{he:weight}. All the Batch Normalization layer weights are initialized to $1$ and are kept in FP32 as HBFP suggests~\cite{drumond:hbfp}.

\input{main/tables/hyperparam_cifar10}

\subsection{Neural machine translation experiments}
\label{app:hyperparams_transformer}
Our implementation is based on the encoder-decoder (seq2seq) Transformer implementation of Fairseq~\cite{ott:fairseq}. We report the hyperparameters we use for all our experiments in Table~\ref{tab:transformerhyperparam}. For the ones we do not specify, we use the default hyperparameters.

\input{main/tables/hyperparam_transformer}

\subsection{Experiments with other transformer models}
\label{app:experiments_transformer}
As mentioned in Section~\ref{sec:experimental_results}, we use the HuggingFace Transformers library~\cite{huggingface:transformers} for BERT-Base training and nanoGPT~\cite{karpathy:nanogpt} for GPT2 training, and Huggingface Timm~\cite{huggingface:timm} for DeiT-Tiny. We use the original hyperparameters mentioned in the respective papers for all these models.

\section{Error bars}
\label{app:error_bars}

\begin{figure}[!ht] 
  \begin{minipage}[b]{\linewidth}
    \centering
    \includegraphics[width=0.47\linewidth]{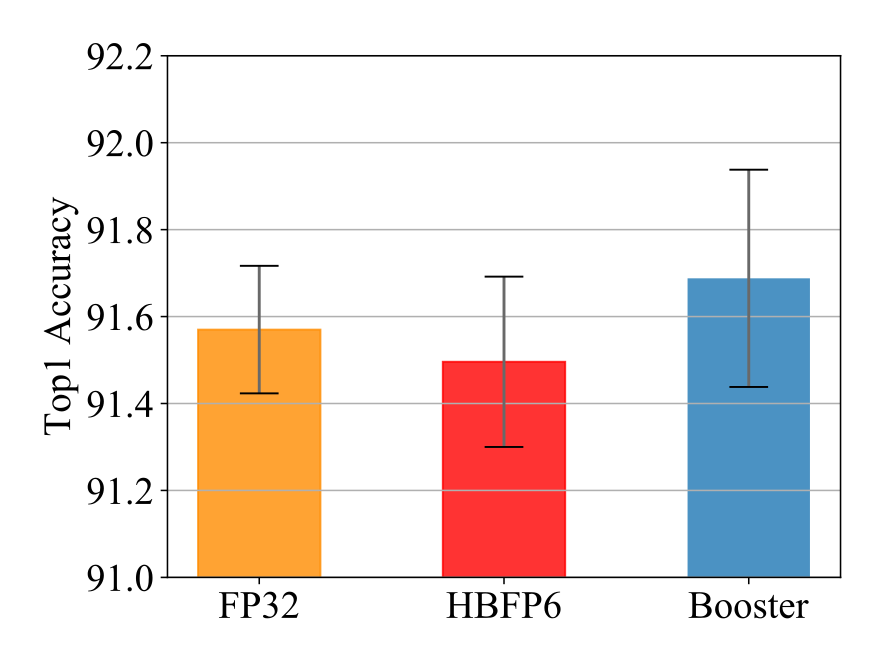} 
  \end{minipage}%%
\caption{Error bars for ResNet20 on CIFAR10 with $5$ different seeds for various configurations.}
\label{fig:error_bars} 
\end{figure}

 In this section, we report the error bars for ResNet20 on CIFAR10 ran on $5$ different seeds for FP32, HBFP$6$, and Accuracy Booster (see Figure~\ref{fig:error_bars}). Columns show the average over the results for the multiple seeds, and the lines indicate the variability of the results (standard deviation). We observe that the variation in results is not more than $0.4\%$. We do not report larger models because multiple runs with such models are computationally expensive.

\section{CNN training curves}
\label{app:curves}
\begin{figure*}[!ht]
\centering
\begin{minipage}{.48\textwidth}
  \centering
  \includegraphics[width=.85\linewidth,height=.7\linewidth]{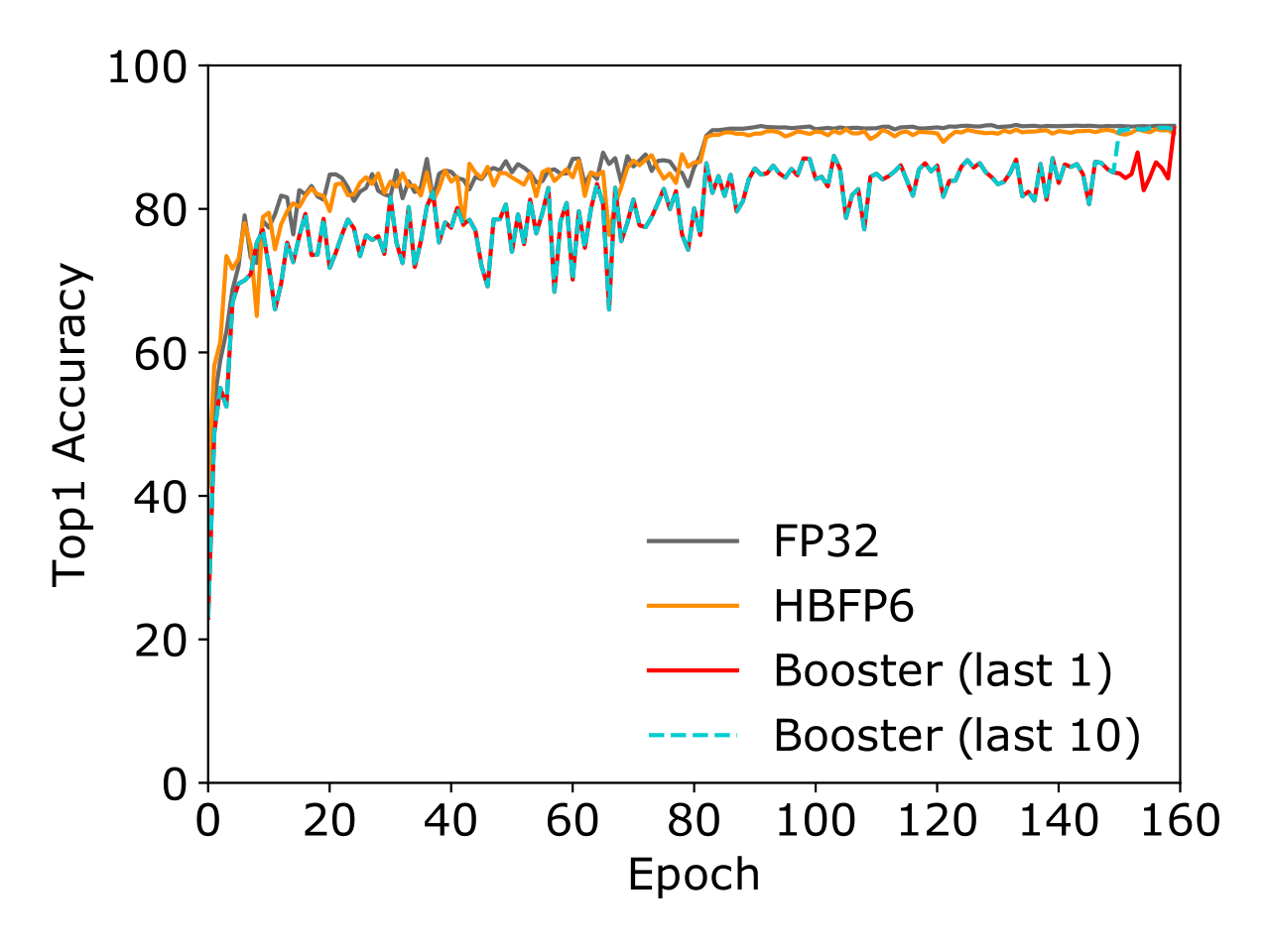}
  % \caption{Top1 Accuracies for ResNet74 on \\ CIFAR100 with various training techniques.}
  % \label{fig:cnn_boosterplots}
\end{minipage}%
\hspace{0.04\textwidth}%
\begin{minipage}{.48\textwidth}
  \centering
  \includegraphics[width=.85\linewidth,height=.7\linewidth]{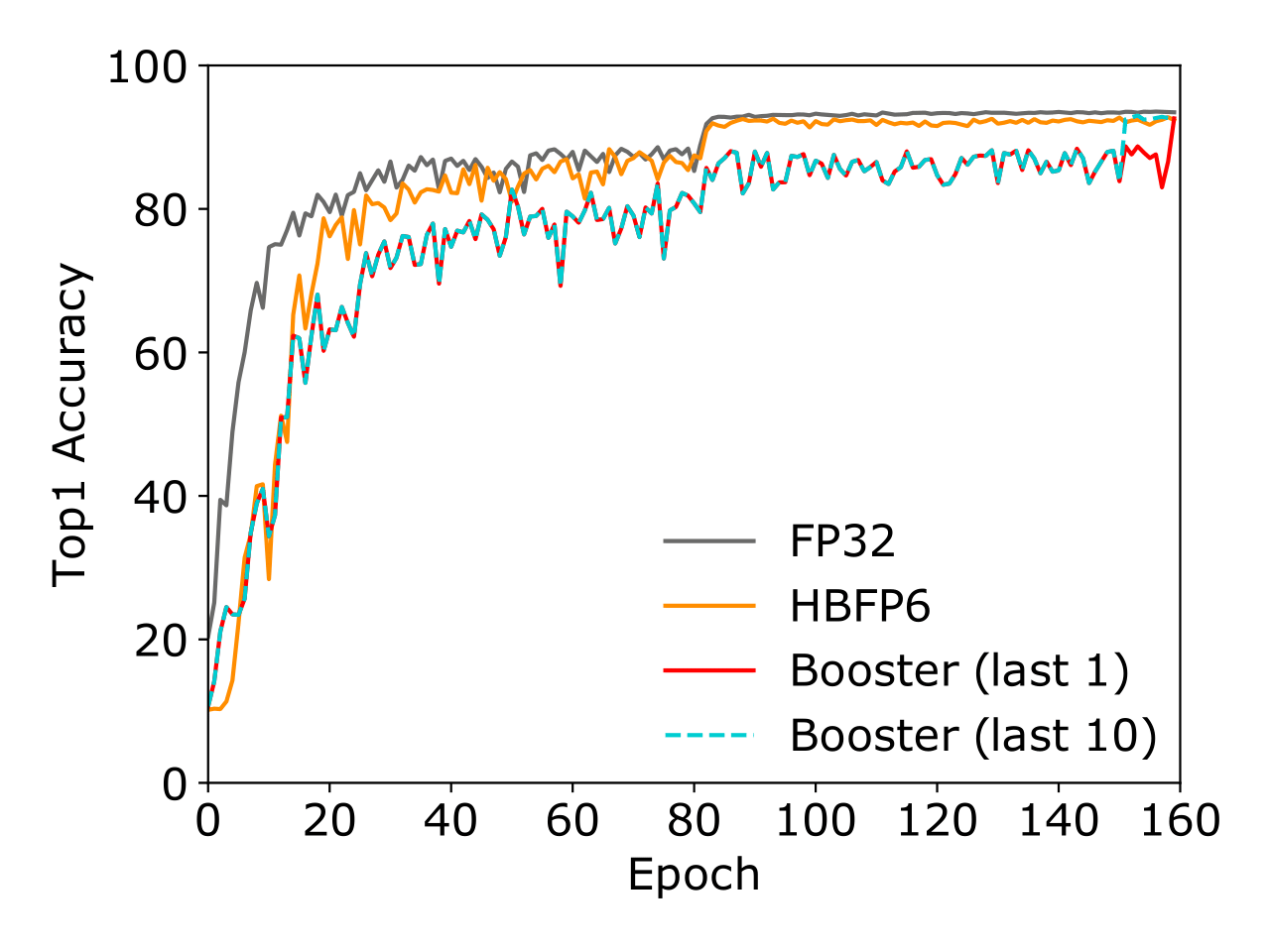}
  % \caption{Top1 Accuracies for DenseNet40 on CIFAR100 with various training techniques.}
  % \label{fig:cnn_boosterplots}
\end{minipage}
\caption{Top1 Accuracy training curves for ResNet20 and ResNet74 on CIFAR10}
\label{fig:cnn_boosterplots1}
\end{figure*}

In this section, we report the training curves of all CNN models trained on CIFAR10 (see Figure~\ref{fig:cnn_boosterplots1}). 

\section{Loss landscapes}
\label{app:loss_landscapes}
\begin{figure*}[t] 
  \begin{minipage}[b]{0.45\linewidth}
    \centering
    \includegraphics[width=\linewidth]{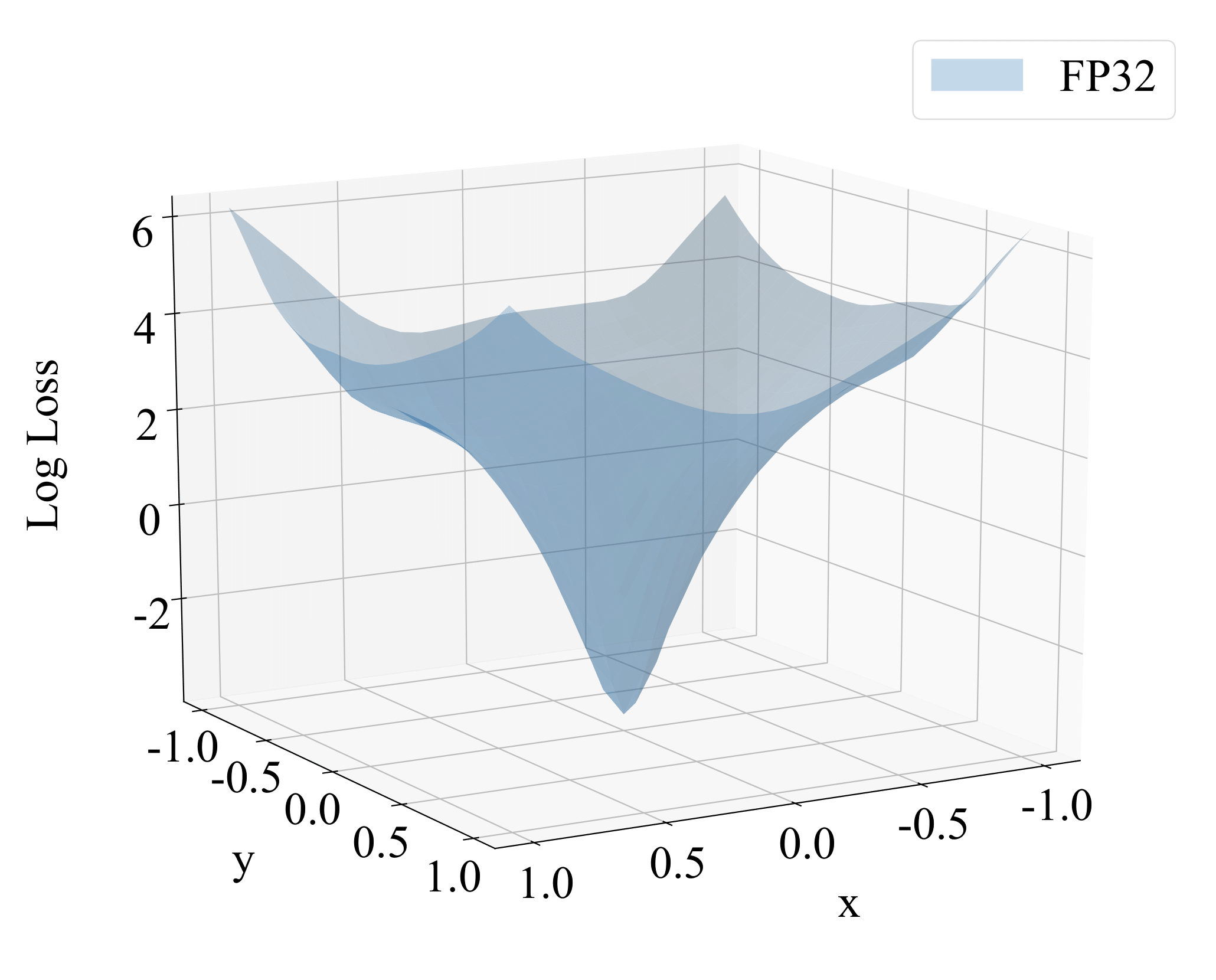} 
    \vspace{-3ex}
  \end{minipage}%%
  \begin{minipage}[b]{0.45\linewidth}
    \centering
    \includegraphics[width=\linewidth]{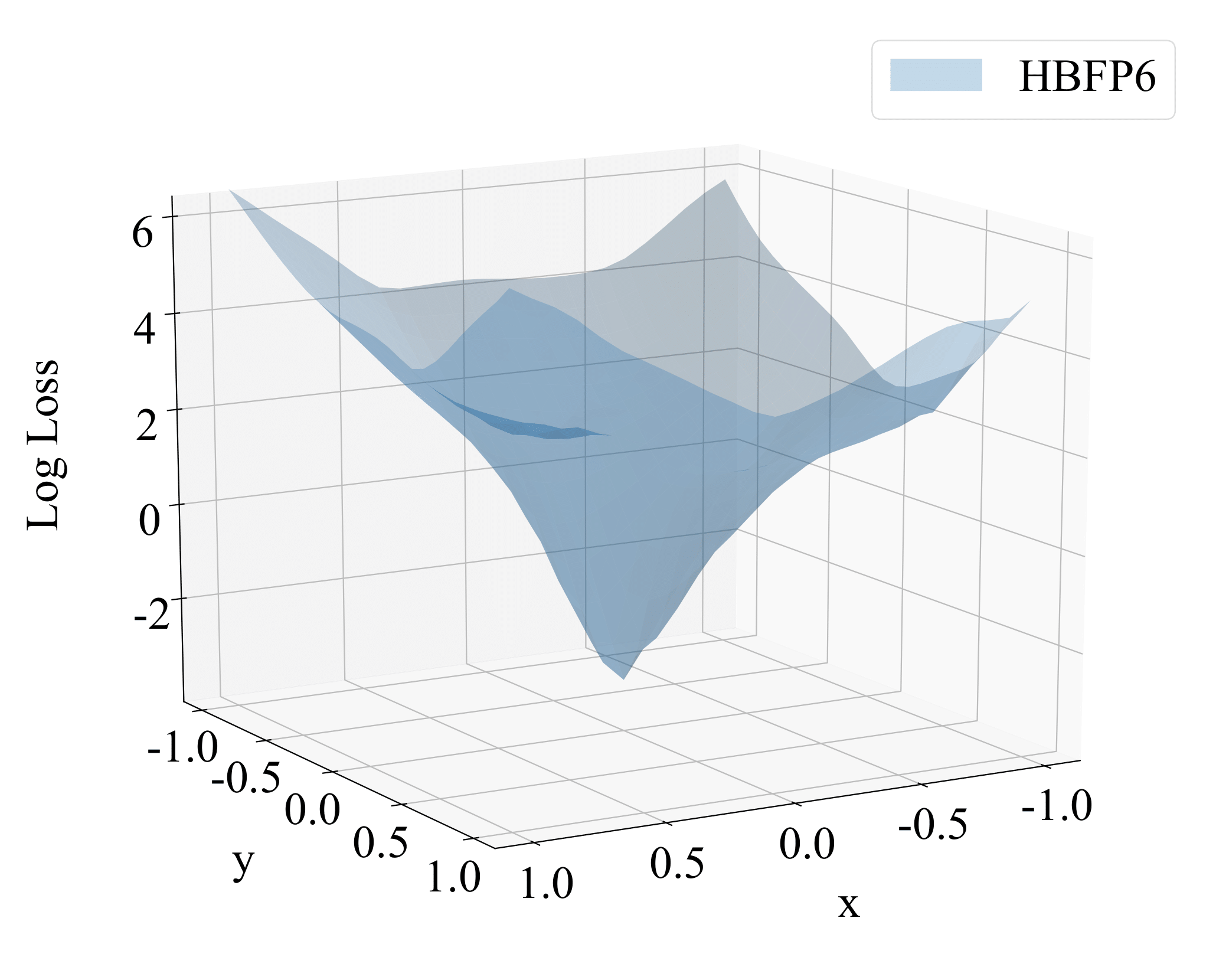} 
    \vspace{-3ex}
  \end{minipage} 
  \begin{minipage}[b]{0.45\linewidth}
    \centering
    \includegraphics[width=\linewidth]{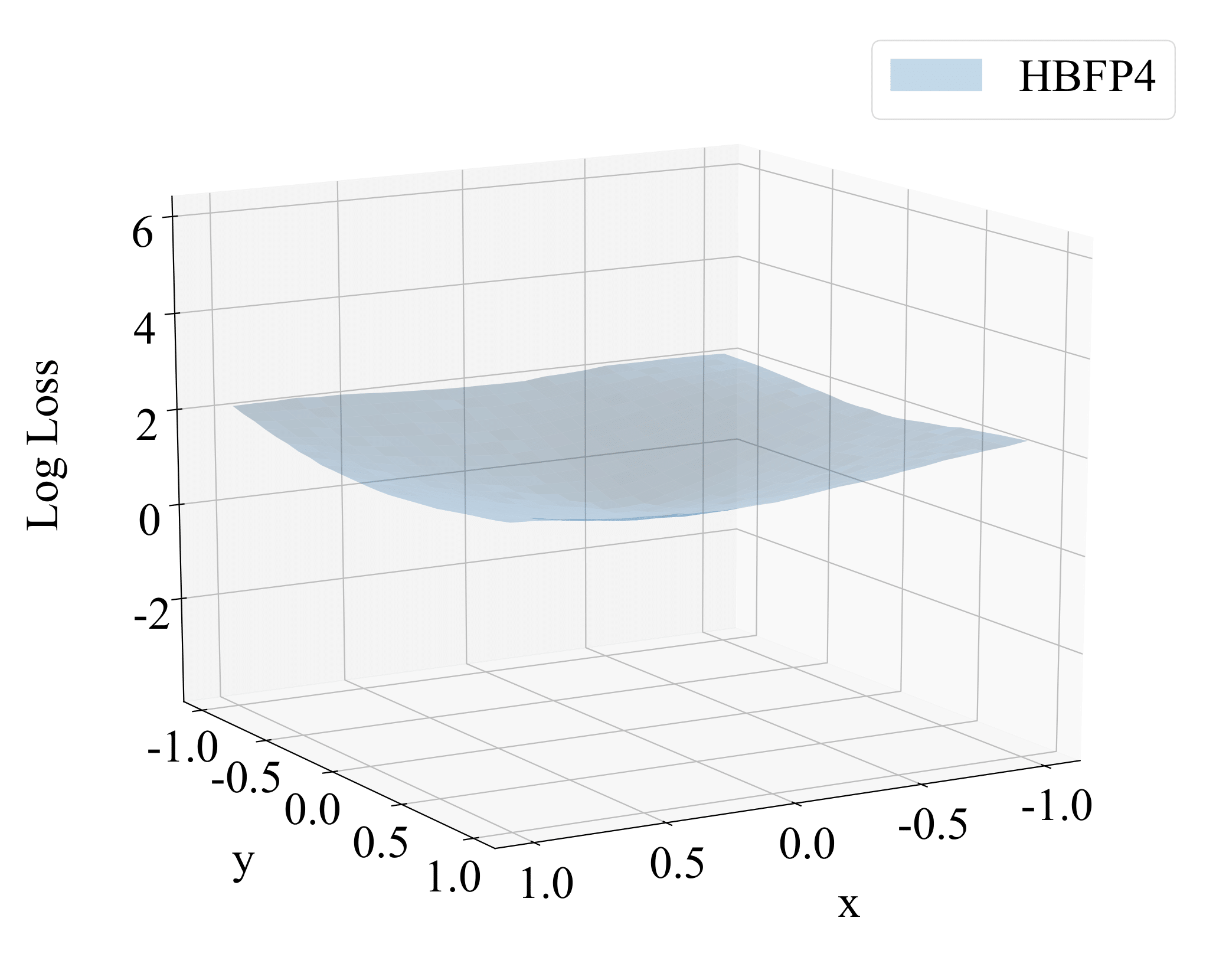} 
    \vspace{-3ex}
  \end{minipage}%% 
  \begin{minipage}[b]{0.45\linewidth}
    \centering
    \includegraphics[width=\linewidth]{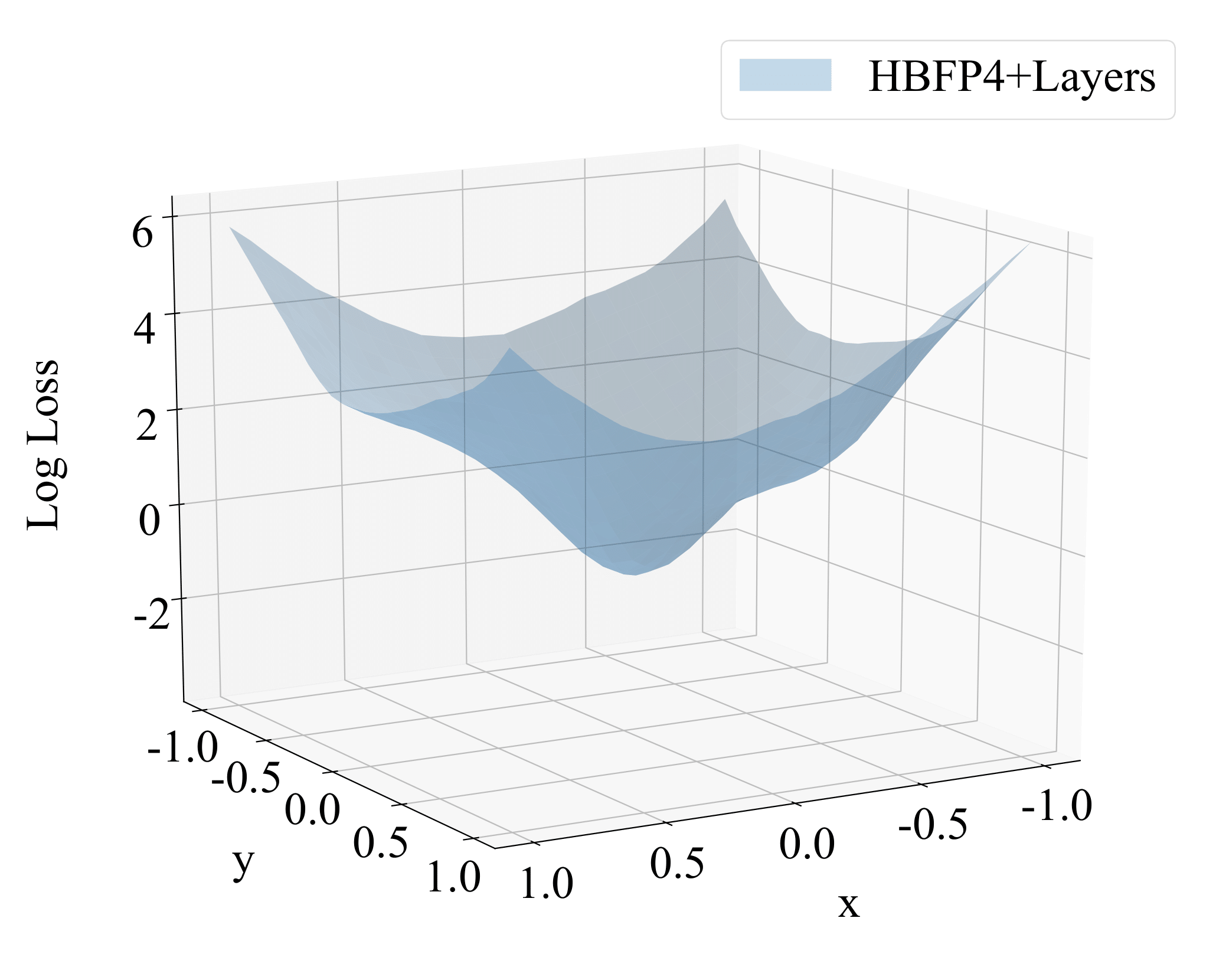} 
    \vspace{-3ex}
  \end{minipage}
  \begin{minipage}{\linewidth}
    \centering
    \includegraphics[width=0.45\linewidth]{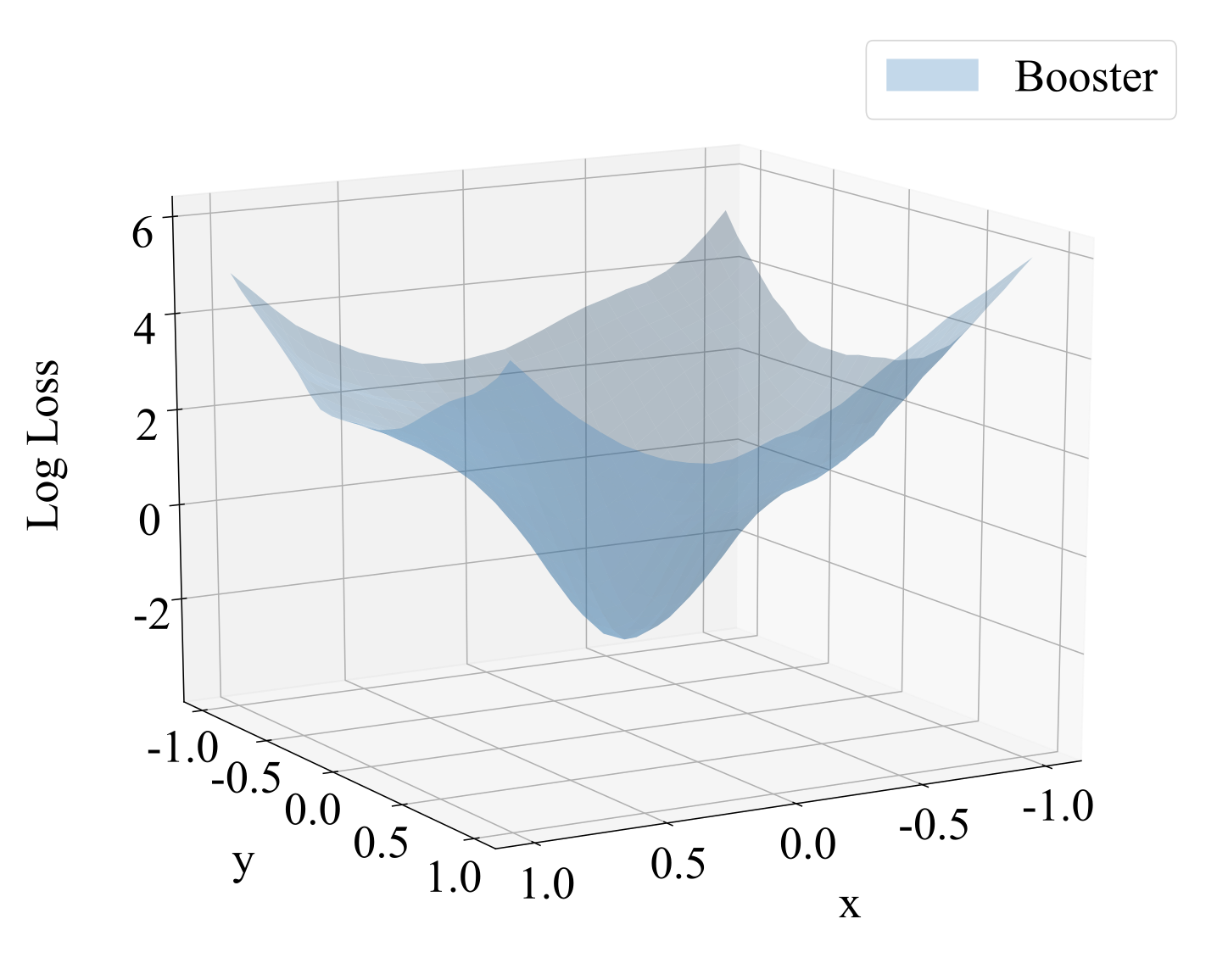} 
    \vspace{4ex}
  \end{minipage}%% 
\caption{Loss landscapes in 3D for ResNet20 on CIFAR10 with various configurations.}
\label{fig:3dplots} 
\end{figure*}

In this section, we report the 3D versions of the loss landscapes explained in Section~\ref{sec:booster}. Figure~\ref{fig:3dplots} shows loss landscapes for ResNet20 trained on CIFAR10 with various configurations.

%% file: main/tables/hbfp5.tex
% \captionsetup{width=0.7\textwidth}
% \captionsetup{justification=centering}
\begin{table*}[h]
  \caption{Top-1 validation accuracies of various CNN models for various HBFP configurations.}
  \small
  \label{tab:hbfp5accuracies}
  \centering
  \resizebox{0.97\columnwidth}{!}{
  \begin{tabular}{ccccccccc}
    \toprule
     % & & \multicolumn{6}{c}{{Models and Datasets}} \\
    % \cmidrule(l{4pt}r{4pt}){3-8}
     & & &
    \multicolumn{2}{c}{CIFAR10} & \multicolumn{4}{c}{CIFAR100}                   \\
    
    \cmidrule(l{4pt}r{4pt}){4-5}
    \cmidrule(l{4pt}r{4pt}){6-9} 
    \makecell{Number\\Format} & Block Size & \makecell{Normalized \\Arithmetic Density}  & ResNet20 & ResNet74 & ResNet50 & ResNet74 & DenseNet40 \\
    \cmidrule(l{3pt}r{3pt}){1-1} 
    \cmidrule(l{3pt}r{3pt}){2-2} 
    \cmidrule(l{3pt}r{3pt}){3-3} 
    \cmidrule(l{3pt}r{3pt}){4-4} 
    \cmidrule(l{3pt}r{3pt}){5-5} 
    \cmidrule(l{3pt}r{3pt}){6-6} 
    \cmidrule(l{3pt}r{3pt}){7-7} 
    \cmidrule(l{3pt}r{3pt}){8-8} 
    \cmidrule(l{3pt}r{3pt}){9-9}

    FP32 & - & $1$ & $91.7$ & $93.6$ & $74.1$ & $74.6$ & $72.4$     \\
    \midrule
    HBFP8 & {576} & $13$ & $91.5$ & $93.4$ & $73.8$ & $74.3$ & $73.7$     \\
    \midrule
    
    \multirow{4}{*}{HBFP$6$} 
     & {$16$} & $15$ & $91.1$ & $93.4$ & $73.1$ & $73.5$ & $72.1$     \\
    % & {$25$} / $17$ & $91.1$ & $92.5$ & $72.6$ & $73.2$ & $71.8$    \\
    % & {$36$} / $17$ & $91.3$ & $92.6$ & $72.5$ & $72.9$ & $71.8$    \\
    % & {$49$} / $18$ & $91.3$ & $92.9$ & $72.3$ & $72.4$ & $71.9$    \\
     & {$64$} & $18$ & $91.1$ & $92.9$ & $72.1$ & $72.4$ & $71.8$    \\
     & {$256$} & $19$ & $91.4$ & $92.8$ & $72.0$ & $72.5$ & $71.5$    \\
     & {$576$} & $20$ & $90.7$ & $92.2$ & $70.8$ & $72.5$ & $71.1$   \\
    \midrule

    \multirow{7}{*}{HBFP$5$} 
     & {$16$} & $17$ & $90.2$ & $92.2$ & $67.2$ & $70.5$ & $69.4$    \\
    % & {$25$} / $19.2$ & $90.2$ & $92.0$ & $67.9$ & $69.0$ & $70.0$     \\
     %& {$36$} / $20.3$ & $90.1$ & $92.0$ & $67.1$ & $69.9$ & $69.0$     \\
     %& {$49$} / $21.1$ & $89.7$ & $92.0$ & $68.7$ & $68.7$ & $69.3$     \\
     & {$64$} & $21$ & $89.7$ & $91.3$ & $66.8$ & $69.3$ & $68.9$     \\
     & {$256$} & $23$ & $89.1$ & $91.1$ & $63.0$ & $66.6$ & $67.8$    \\
     & {$576$} & $23$ & $88.9$ & $91.0$ & $64.6$ & $64.4$ & $67.6$     \\
    \midrule
    
    % \multirow{7}{*}{\makecell{HBFP$5$ with \\ first and last\\layers in FP32}} 
    %  & {16} / $-$ & $90.80$ & $92.56$ & $71.23$ & $72.31$ & $70.90$     \\
    %  & {25} / $-$ & $90.62$ & $91.79$ & $71.76$ & $72.58$ & $70.52$     \\
    %  & {36} / $-$ & $90.62$ & $92.73$ & $71.45$ & $72.26$ & $70.01$     \\
    %  & {49} / $-$ & $90.49$ & $92.02$ & $70.51$ & $72.07$ & $70.16$     \\
    %  & {64} / $-$ & $90.31$ & $92.41$ & $69.59$ & $72.09$ & $70.10$    \\
    %  & {256} / $-$ & $90.07$ & $92.11$ & $68.76$ & $70.20$ & $69.21$    \\
    %  & {576} / $-$ & $90.37$ & $91.06$ & $68.66$ & $69.14$ & $68.78$     \\
    % \midrule
        
    \multirow{4}{*}{HBFP$4$} 
     & {$16$} & $20$ & $82.6$ & $76.9$ & - & - & $63.7$    \\
   %  & {$25$} / $22$ & $81.8$ & $78.6$ & - & - & $64.3$     \\
   %  & {$36$} / $23$ & $80.8$ & $76.6$ & - & - & $63.3$     \\
   %  & {$49$} / $24$ & $79.3$ & $71.2$ & - & - & $65.6$     \\
     & {$64$} & $25$ & $80.2$ & $74.4$ & - & - & $62.4$    \\
     & {$256$} & $27$ & $77.0$ & $60.7$ & - & - & $60.0$   \\
     & {$576$} & $27$ & $75.3$ & $66.7$ & - & - & $59.8$     \\
    \midrule\midrule

    \multicolumn{3}{c}{{\makecell{Total Number of FLOPs \\ required to train the model}}} & $41$M & $174$M & $119$M & $326$M & $542$M \\   
    %\multicolumn{3}{c}{{Number of Model Parameters}} & $270$K & $38$M & $25$M & $38$M & $1$M     \\
   
    \bottomrule
  \end{tabular}
  }
\end{table*}

%% file: main/tables/hyperparam_cifar10.tex
\begin{table}[!ht]
  \caption{Hyperparameters for various sizes of ResNet on CIFAR10 and CIFAR100}
  \label{tab:cnnhyperparam}
  \centering
  
  \resizebox{0.57\columnwidth}{!}{
  \begin{tabular}{lcc}
    \toprule
    Hyperparameter & \makecell{Resnet20 \& \\ ResNet74 \\ on CIFAR10} & \makecell{DenseNet40 \& \\ ResNet74\\ on CIFAR100}\\
    \toprule
    
    Epochs & $160$ & $300$\\
    Batch size & $128$ & $128$\\
    Optimizer & SGD & SGD\\
    Nesterov & True & True\\
    Learning rate & $0.1$ & $0.1$\\
    Learning rate decay epochs & $82,122$ & $150,225$\\
    Learning rate warmup & True & True\\
    Weight decay & $1e$-$04$ & $1e$-$04$\\
    Momentum & $0.9$ & $0.9$\\

    \bottomrule
  \end{tabular}}
\end{table}

%% file: main/tables/hyperparam_transformer.tex
\begin{table}[!ht]
  \caption{Hyperparameters for Transformer on IWSLT'14 German to English translation task}
  \label{tab:transformerhyperparam}
  \centering
  
  \resizebox{0.62\columnwidth}{!}{
  \begin{tabular}{ll}
    \toprule
    \multicolumn{2}{c}{Transformer architecture hyperparameters} \\
    \toprule
    
    Dataset & iwslt14.tokenized.de-en\\
    Architecture & transformer\_iwslt\_de\_en\\
    Number of layers & $6$\\
    Number of heads & $4$\\
    Number of hidden dimensions & $512$\\
    Embedding dimension for FFNs & $1024$\\
    Optimizer & Adam\\
    Adam-betas & $(0.9, 0.98)$\\
    Learning rate & $0.0005$\\
    Learning rate scheduler & Inverse Square Root\\
    Warmup updates & $4000$\\
    Warmup initial learning rate & $1e$-$07$\\
    Label smoothing & $0.1$\\
    Maximum tokens & $4096$\\
    Criterion & label\_smoothed\_cross\_entropy\\
    Clip-norm & $0.0$\\
    Weight decay & $0.0001$\\
    Dropout & $0.3$\\

    \bottomrule
  \end{tabular}}
\end{table}